\pdfoutput=1
\documentclass[journal]{IEEEtran}

\usepackage[utf8]{inputenc}

\usepackage{amsmath}
\usepackage{amssymb}
\usepackage{amsfonts}
\usepackage{mathtools}
\usepackage{comment}
\usepackage{balance}
\usepackage{cite}
\usepackage{subcaption}
\usepackage{caption}
\usepackage{relsize}
\usepackage{multirow}
\usepackage{adjustbox}
\usepackage{nccmath}

\DeclareMathOperator*{\argmin}{arg\,min}

\newcommand*{\horzbar}{\rule[.5ex]{2.5ex}{0.5pt}}

\usepackage{algorithm}
\usepackage{algorithmic}
\usepackage{lipsum}
\begin{document}

\title{Modeling of Spatio-Temporal Hawkes Processes with Randomized Kernels}

\author{Fatih~Ilhan and~Suleyman~S.~Kozat,~\IEEEmembership{Senior Member,~IEEE}
\thanks{This work in by part supported by Turkish Academy of Sciences Outstanding Researcher program.}
\thanks{F. Ilhan and S. S. Kozat are with the Department of Electrical and Electronics Engineering, Bilkent University, Ankara 06800, Turkey, e-mail: filhan@ee.bilkent.edu.tr, kozat@ee.bilkent.edu.tr.}
\thanks{F. Ilhan and S. S. Kozat are also with DataBoss A. S., Bilkent Cyberpark, Ankara 06800, Turkey, email: fatih.ilhan@data-boss.com.tr, serdar.kozat@data-boss.com.tr.}
}

\maketitle

\begin{abstract}

We investigate spatio-temporal event analysis using point processes. Inferring the dynamics of event sequences spatio-temporally has many practical applications including crime prediction, social media analysis, and traffic forecasting. In particular, we focus on spatio-temporal Hawkes processes that are commonly used due to their capability to capture excitations between event occurrences. We introduce a novel inference framework based on randomized transformations and gradient descent to learn the process. We replace the spatial kernel calculations by randomized Fourier feature-based transformations. The introduced randomization by this representation provides flexibility while modeling the spatial excitation between events. Moreover, the system described by the process is expressed within closed-form in terms of scalable matrix operations. During the optimization, we use maximum likelihood estimation approach and gradient descent while properly handling positivity and orthonormality constraints. The experiment results show the improvements achieved by the introduced method in terms of fitting capability in synthetic and real-life datasets with respect to the conventional inference methods in the spatio-temporal Hawkes process literature. We also analyze the triggering interactions between event types and how their dynamics change in space and time through the interpretation of learned parameters. 
\end{abstract}

\begin{IEEEkeywords}
Parameter estimation, time series, system modeling, Point processes, random Fourier features, event analysis.
\end{IEEEkeywords}

\IEEEpeerreviewmaketitle

\section{Introduction}

\label{sec:intro}

\subsection{Preliminaries}

\IEEEPARstart{W}{e study} spatio-temporal event analysis using point processes, which has several applications in signal processing, computer networks, security and forecasting applications~\cite{tsp_social,crowd_mpp,tsp_poisson,tsp_wireless,meme}. Most of the real-world events exhibit certain spatio-temporal patterns such as correlation, causation, and excitation, which can be modeled as a system whose latent structure is reflected into real-world with their realizations. Modeling and learning this structure is important due to its promising applications such as network analysis, event prediction and hotspot detection~\cite{st_pattern,mohler_crime,tsp_sparsity,tsp_gaussian,tsp_seismic,yuan_mvhawkes,tnnls_driver,ogata_earthquake}. In this context, we analyze the triggering relations between events in a given sequence, and how these interactions evolve in space and time, which can be useful for forecasting and policy planning for security and business applications~\cite{mohler_crime, ogata_earthquake}. To this end, we model the event sequence using point processes, which directly represent the underlying structure of spatio-temporal excitations by their internal parameters. Therefore, inferring these parameters provides an interpretable and forthright way to analyze the given spatio-temporal data.

Point processes are used to capture the dynamics of the event sequence by expressing their rate of occurrences with an intensity function conditioned on the history~\cite{ppbook}. In our problem, events are described by their locations, times and types. Therefore, we consider a multi-dimensional form of point processes called as spatio-temporal point processes~\cite{yuan_mvhawkes,stpp_emergency,review}. We particularly work on spatio-temporal Hawkes processes that have a self-exciting nature by their default form, in which the intensity value is triggered by past events. In this approach, the excitation between events is usually modeled as to be decaying exponentially in time with an exponential kernel, and in space with a Gaussian kernel~\cite{review, yuan_mvhawkes}. 

Although modeling of temporal excitation with exponential decay is shown to be effective in most scenarios, the assumption that spatial excitation can be completely represented with a Gaussian kernel may not hold in all cases. Hence, we introduce certain degree of randomness to the spatial kernel by using randomized kernel representation. We utilize random Fourier features~\cite{rahimi_rff}, which approximate the output of a shift-invariant continuous kernel using the inner products of the embedded vectors. Flexing the structure of the spatial kernel enables our model to capture excitation without purely Gaussian decay in the spatial domain. The number of dimensions in the transformed feature space is a hyperparameter, which directly controls the randomization effect, thus we can readily tune it depending on the spatial characteristics of the given event sequence. In addition, replacing the pairwise kernel calculations with randomized vector products enables us to formulate the problem in a neat matrix form, which increases the scalability of the introduced framework.

We optimize the parameters of the process using maximum likelihood estimation (MLE) approach in terms of negative log-likelihood, which is shown to be quite efficient, consistent and asymptotically unbiased for point processes~\cite{review}. Therefore, we define our evaluation metric in terms of negative log-likelihood per event, which directly expresses the fitting performance. In order to learn the parameters of a spatio-temporal Hawkes process, there exist several inference methods in the point process literature, most notably, expectation-maximization (EM)~\cite{yuan_mvhawkes,latent_spatnet,em_inference} algorithm and stochastic declustering~\cite{zhuang_stodecl,chiodi2015,Marsan2010,fox2016}. Recently, gradient descent-based optimization methods has also been preferred in the context of temporal point processes~\cite{rein_tp,nandu,wang2018graph}. Nevertheless, employing gradient descent in the spatio-temporal case is not as straightforward as in the temporal case. The difficulty lies within the structure of the likelihood function, which includes multi-dimensional integrals of kernel outputs. Hence, maximum likelihood estimation with gradient-based methods is not directly viable~\cite{review}. The conditional intensity function of a temporal point process is only defined along the temporal dimension, hence expressing likelihood objective in a differentiable manner, and applying gradient descent-based optimization is rather straightforward compared to the spatio-temporal case. To address this issue, we analytically derive the intractable terms in the likelihood function and their derivatives. We also employ reparameterization techniques and projected gradient-descent to handle the numerical constraints over the process parameters properly.

Even though there exists a considerable amount of prior art about spatio-temporal Hawkes processes, we, for the first time in the literature, utilize random Fourier features based kernel representation for spatio-temporal Hawkes processes. Our approach provides flexibility thanks to the introduced controllable randomization over spatial modeling. We also introduce a novel inference framework with well-organized matrix formulations and gradient descent-based optimization, which provides scalability. Furthermore, we investigate the fitting performance of the introduced method through an extensive set of experiments involving synthetic and real-life datasets. The results show that our method provides significant improvements compared to the EM algorithm and stochastic declustering, which are commonly favored in the point process literature~\cite{yuan_mvhawkes,latent_spatnet,em_inference,zhuang_stodecl,chiodi2011,Marsan2008}. Finally, we perform event analysis over real-life datasets by interpreting the inferred parameters.

\subsection{Prior Art and Comparisons}

A significant amount of research has been conducted in signal processing, applied mathematics, and machine learning literatures to learn and apply spatio-temporal point processes~\cite{tsp_social,tsp_wireless,mohler_crime,yuan_mvhawkes}. The approach of spatio-temporal modeling with point processes has been applied to various real-world scenarios such as seismological modeling of earthquakes and aftershocks~\cite{ogata_earthquake,verejones,chiodi2011,Marsan2008}, criminological modeling of the dynamics of illegal incidents~\cite{mohler_crime,wang2018graph}, forecasting of disease outbreaks~\cite{review}, network analysis~\cite{yuan_mvhawkes,netest_ieeeit,latent_icml,recip_nips,latent_spatnet}, and so on. When carefully analyzed, the behavior of underlying systems can vary among different contexts. To this end, several forms of point processes have been proposed with different characteristics, such as Poisson process, Cox process, self-correcting processes~\cite{self_correct}, and self-exciting processes~\cite{ppbook}. In this study, we consider spatio-temporal Hawkes process, which was first applied for earthquake prediction~\cite{ogata_earthquake} and then successfully adapted to other applications such as crime analysis~\cite{mohler_crime}. 

While modeling spatio-temporal Hawkes processes, there has been several proposals for the form of spatial kernel such as isotropic kernels~\cite{ogata_earthquake}, diffusion kernels~\cite{verejones}, and Gaussian kernels~\cite{mohler_crime, yuan_mvhawkes}. Even though the proposed forms have the common characteristic of having an inverse relation between the excitation level and distance from the event center, their behaviors are considerably different. On the contrary, even though we employ Gaussian kernel as well, we introduce randomization to the spatial modeling of the problem through random Fourier features based kernel representation. The introduced tunable randomization while modeling the spatial excitation enhances the performance in real-life scenarios, particularly when the spatial dynamics of the underlying system deviates from pure Gaussian behavior.

Random Fourier features have been used successfully to increase the scalability of kernel-based methods such as support vector machines (SVMs)~\cite{online_rff}. Introducing randomization to the learning process has also been studied in machine learning literature~\cite{huang_elm}. From the neural network perspective, our representation can be interpreted as a perceptron layer with randomly initialized weights and sinusoidal activation function, where the weights are not being updated and they are sampled from a distribution related to the spectral distribution of the kernel~\cite{rffelm}. This architecture is called as extreme learning machines (ELM) that have universal approximation capability as the number of nodes (embedding dimensions) goes to infinity~\cite{huang_elm}. Since we only have to approximate the kernel outputs of two-dimensional spatial vectors, low embedding dimensions suffice with negligible approximation errors~\cite{rahimi_rff}. This enables us to replace complicated pairwise kernel calculations with scalable matrix operations.

To increase the generalization power in point process models, machine learning based approaches have also been proposed in recent studies~\cite{nandu,mai_neural_hawkes,wang2018graph,rein_tp}. In particular, recurrent neural networks (RNNs) and variants such as long short-term memory networks (LSTMs) are employed in the context of temporal point processes~\cite{nandu,mai_neural_hawkes}. However in the spatio-temporal domain, increased number of dimensions and sparsity may lead to unstable and difficult training of machine learning models~\cite{sparse}. In \cite{mai_neural_hawkes}, authors employ LSTMs to model temporal Hawkes processes and uses the Monte-Carlo estimation of the intractable terms in the likelihood function and parameter gradients. However, relying on Monte-Carlo estimation is also problematic due to the increased space size after introducing spatial dimensions, which would require a large number of samples and result in degraded approximation performance~\cite{mc_fail}. Since we formulate the problem in a tractable form, our approach does not need any sampling based approximation.

In addition to the application and modeling perspectives, there is also an extensive literature about estimating the parameters of a spatio-temporal Hawkes process. Several inference methods were proposed for this problem. The most commonly used techniques involve MLE approach, which can be solved using expectation maximization (EM) as applied in various studies~\cite{yuan_mvhawkes,latent_spatnet,em_inference}. EM algorithm exhibits certain nice properties such as consistently increasing likelihood at each iteration, and naturally producing valid estimations for desired parameters without any numerical constraints~\cite{xu_em_convergence}. However, it can suffer from instability due to bad initialization and slow convergence in regions, where the likelihood function is flat~\cite{couvreur_em_tour}. 

Another method, stochastic declustering has shown successful results particularly for earthquake modeling~\cite{zhuang_stodecl,chiodi2011,chiodi2015,Marsan2008,Marsan2010,fox2016}. This is a non-parametric approach and relies on the branching structure of events, which assumes that the events can be clustered into two separate groups: background events and triggered events branching from the background. Bayesian inference methods have also been studied for self-exciting temporal point processes~\cite{latent_icml,recip_nips,Ogata1988,rasmussen_bayes}. Finally, gradient descent-based numerical optimization methods are used in the most recent studies~\cite{mai_neural_hawkes,nandu,wang2018graph}. In this study, we optimize the parameters through likelihood maximization with gradient descent-based algorithms since these methods are shown to be simple yet effective, particularly for neural networks~\cite{tsp_chaotic,rein_tp}.

\subsection{Contributions}

Our main contributions are as follows:

\begin{enumerate}
    \item As the first time in the literature, we apply random Fourier features based transformations to represent kernel operations in spatio-temporal Hawkes processes. This transformation increases the flexibility of our spatial modeling due to the introduced randomization, and can easily be controlled by tuning the number of embedding dimensions depending on the application.
    \item We introduce a novel framework to formulate the problem in terms of scalable matrix operations by utilizing the vector products of transformed features instead of explicit pairwise kernel calculations.
    \item We employ gradient descent based optimization to learn the parameters of the proposed model. To this end, we analytically obtain the intractable terms of the likelihood and properly handle the constraints over parameters by using reparameterization techniques and projected gradient descent.
    \item We propose a simulation algorithm that follows thinning procedure to generate synthetic spatio-temporal Hawkes process realizations with multiple event types.
    \item Through an extensive set of experiments over synthetic and real-life datasets, we demonstrate that our method brings significant improvements in terms of fitting performance with respect to the EM algorithm and stochastic declustering, which have been extensively favored in the point process literature~\cite{review,em_inference,yuan_mvhawkes,zhuang_stodecl,chiodi2011,Marsan2008}.
    \item We demonstrate the practical applications of the proposed method by performing event analysis over real-life spatio-temporal event sequences through the interpretation of inferred excitation coefficients.
\end{enumerate}

\subsection{Organization}

The remainder of the paper is organized as follows. We provide the form of the spatio-temporal Hawkes process and introduce the optimization problem in Section~\ref{sec:problem}. Then we provide the matrix formulations to express the likelihood function in a closed-form using random Fourier features in Section~\ref{sec:method_matrix}. Then, we analytically obtain the derivatives of the likelihood with respect to process parameters and provide in Appendix~\ref{sec:app_a}. In Section~\ref{sec:method_algorithm}, we give the gradient-based optimization algorithm for maximum likelihood estimation under parameter constraints. We analyze the fitting performance of the proposed method over simulated and real-life datasets and perform network analysis in Section~\ref{sec:experiments}. We conclude the paper in Section~\ref{sec:conclusion} with several remarks.

\section{Problem Description}

\label{sec:problem}

\begin{figure}[t]
    \centering
    \includegraphics[width=\columnwidth]{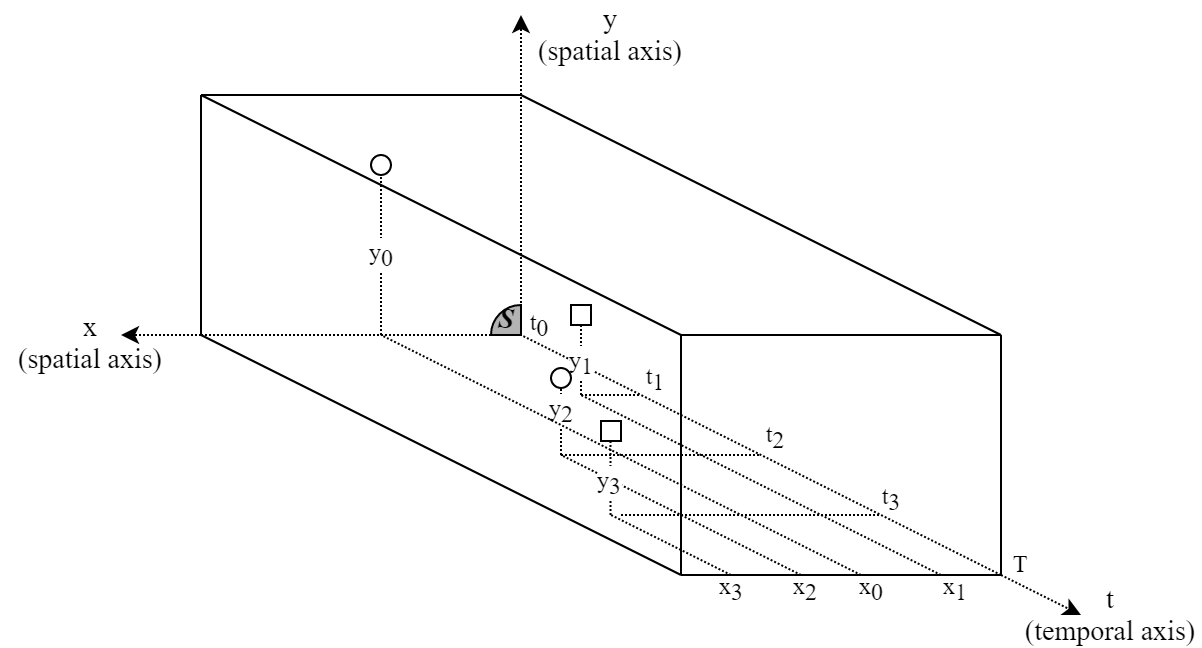}
    \caption{An example of a spatio-temporal event sequence with four events. Each event is located inside the spatial domain $\mathcal{S}$, and distributed along the temporal axis. We use various shapes to represent different event types. Here, we have two types of events.}
    \label{fig:events}
\end{figure}

In this paper\footnote{All vectors are column vectors and denoted by boldface lower case letters. Matrices are denoted by boldface upper case letters. $\boldsymbol{x}^T$ and $\mathbf{X}^T$ are the corresponding transposes of $\boldsymbol{x}$ and $\mathbf{X}$. $\|\boldsymbol{x}\|$ is the $\ell^2$-norm of $\boldsymbol{x}$. $\odot$ and $\oslash$ denotes the Hadamard product and division operations. $|\mathbf{X}|$ is the determinant of $\mathbf{X}$. For any vector $\boldsymbol{x}$, $x_i$ is the $i^{th}$ element of the vector. $x_{ij}$ is the element that belongs to $\mathbf{X}$ at the $i^{th}$ row and the $j^{th}$ column. $\text{sum}(\cdot)$ is the operation that sums the elements of a given vector or matrix. $\delta_{ij}$ is the Kronecker delta, which is equal to one if $i=j$ and zero otherwise.}, we study spatio-temporal event analysis with point processes. We observe an event sequence $\mathcal{E}=\{e_i\}_{i=1}^{N}$, and model it with spatio-temporal Hawkes processes. Here, $N$ is the total number of events and $e_i=\{u_i, t_i, \boldsymbol{s_i}\}_{i=1}^{N}$ is the $i^{th}$ event in the sequence with type $u_i \in \mathbb{N}$, time $t_i \in \mathcal{T}$ and location $\boldsymbol{s_i}=[x_i, y_i]^T \in \mathcal{S}$.\footnote{We define temporal space $\mathcal{T} \triangleq \{t \; | \; t \in [0, \infty)\}$, and consider spatial space $\mathcal{S}$ to be a rectangular subset of $\mathbb{R}^2$.} We visualize an example of a spatio-temporal event sequence in Fig. ~\ref{fig:events}. Our goal is to infer the parameters of the process and perform analysis on the given event sequence through investigating the excitation between events in spatial and temporal domain. We model the spatial dynamics of $\mathcal{E}$ by expressing the kernels with random Fourier features-based kernel representations. The parameters of the process are denoted as $\boldsymbol{\theta}$, and we optimize them using MLE approach, in which the objective function is the log-likelihood of the given event sequence. Then, we solve the underlying optimization problem with gradient descent and properly handle numerical constraints using reparameterization techniques and projected gradient descent.

\subsection{Temporal Point Processes}

A temporal point process is a stochastic process that consists of realizations of subsequent events in discrete time $t_i \in \mathbb{R}$ with $i \in \mathbb{Z}$. We can interpret a temporal point process by specifying the distribution of the time distance between subsequent events (inter-event times). Let $f^*(t)=f(t|\mathcal{H}_t)$ be the conditional density function for the time of the next event given the time history of events $\mathcal{H}_t$. To express the past dependence in an evolutionary point process, conditional intensity function is defined as follows~\cite{review,ppbook},
\noindent
\begin{equation}\label{eq:lambda}
    \lambda^*(t) = \frac{f^*(t)}{1-F^*(t)},
\end{equation}
\noindent
where $F^*(t)$ is the cumulative density distribution of $t$ such that $F^*(t) = 1 - \exp{\left(-\int_{t_n}^t \lambda^*(t)\right)}$ and $t_n$ is the time of the last event before $t$. We can express the conditional density function $f^*(t)$ in terms of conditional intensity function $\lambda^*(t)$ using \eqref{eq:lambda} as
\noindent
\begin{equation} \label{eq:density}
    f^*(t) = \lambda^*(t)e^{-\Lambda_{\lambda^*}(t)}, \quad \Lambda_{\lambda^*}(t) = \int_{t_n}^t{\lambda^*(\tau)d\tau},
\end{equation}
\noindent
where $t_n$ is the time of the last event before $t$. Here, the conditional intensity function can have many forms. As a simple example, in the case of a Poisson process, $\lambda^*(t)=\lambda(t)=\lambda$, i.e value of the conditional intensity function is constant through time. 

\subsection{Hawkes Processes}

Unlike the Poisson Process, Hawkes process has an evolutionary nature, in which the events excite each other depending on their types and distance as expressed in the following form:
\noindent
\begin{equation}
    \lambda_u^*(t) = \mu_u + \sum_{j|t_j<t}k_{u_ju}g(t, t_j, u, u_j), \nonumber
\end{equation}
\noindent
where  $\mu_{u}$ denotes the background conditional intensity, $k_{u_ju}$ is the excitation of event type $u_j$ over $u$ for triggering conditional intensity and $g(t, t_j, u, u_j)$ is the output of the temporal triggering kernel evaluated at event times $t$ and $t_j$ for given event types $u$ and $u_j$. This form enables us to model the point processes that show temporally clustered patterns.

\subsection{Spatio-Temporal Hawkes Processes}

In the spatio-temporal case, each event also has a spatial vector ($\boldsymbol{s}$) that describes its location. While expressing the conditional intensity function, we consider the following form in our problem\footnote{Note that ${}^*$ sign, which denotes the conditionality on history will be dropped from now on for the sake of notational simplicity.}:
\noindent
\begin{equation}
    \lambda_{u}(t, \boldsymbol{s}) = \mu_{u}(\boldsymbol{s}) + \gamma_{u}(t, \boldsymbol{s}), \label{eq:lambda_generic}
\end{equation}
\noindent
where $\mu_{u}(\boldsymbol{s})$ denotes the base conditional intensity for spatial vector $\boldsymbol{s}$ and event type $u$, and $\gamma_{u}(t, \boldsymbol{s})$ denotes the triggering conditional intensity for any time $t$, $\boldsymbol{s}$ and $u$. We can parametrize the base and triggering conditional intensities in \eqref{eq:lambda_generic} as follows,
\noindent
\begin{align}
    & \mu_{u}(\boldsymbol{s}) = \frac{1}{T}\sum_{j=1}^N k_{u_ju}^{(\mu)} g_2^{(\mu)}(\boldsymbol{s}, \boldsymbol{s_j}), \label{eq:mu_intro} \\
    & \gamma_{u}(t, \boldsymbol{s}) = \sum_{j|t_j<t}k_{u_ju}^{(\gamma)}g_1(t, t_j,u,u_j)g_2^{(\gamma)}(\boldsymbol{s}, \boldsymbol{s_j}), \label{eq:gamma_intro}
\end{align}
\noindent
where $g_1$ is the temporal kernel function and $g_2^{(.)}$ is the spatial kernel function.\footnote{For any scalar, vector, matrix or function, we denote the belonging to the intensity component ${(\cdot)}$ with power notation, e.g, $g_2^{(\mu)}$ is the spatial kernel parameterized for base intensity component.} These functions can be expressed as
\noindent
\begin{align}
    & g_1(t, t_j, u, u_j) = w_{u_ju}e^{-w_{u_ju}(t-t_j)}
\end{align}
\noindent
and
\noindent
\begin{align}
    & g_2^{(\cdot)}(\boldsymbol{s}, \boldsymbol{s_j}) = \frac{1}{2\pi} |\mathbf{\Sigma}^{(\cdot)}|^{-1/2} e^{-\frac{1}{2}(\boldsymbol{s}-\boldsymbol{s_j})^T \mathbf{\Sigma^{(\cdot)}}^{-1} (\boldsymbol{s}-\boldsymbol{s_j})}, \label{eq:g2}
\end{align}
\noindent
where $T$ is the duration of the event sequence, $N$ is the number of events, $\mathbf{\Sigma}^{(\cdot)}$ is the covariance matrix of the spatial Gaussian kernel, and $w_{u_j u} \geq 0$ is the decay rate of the intensity triggered by event type $u_j$ over $u$. 

The excitation values ($k_{ij}$) and weight decays ($w_{ij}$) are expressed in form of matrices $\mathbf{K}$ and $\mathbf{W}$, where $k_{ij}, w_{ij} > 0$. The multivariate normal distribution is said to be non-degenerate when the symmetric covariance matrix $\mathbf{\Sigma}^{(.)}$ is positive definite. In this case, $g_2^{(.)}(\boldsymbol{s}, \boldsymbol{s_j})$ will have an invertible covariance matrix and density.

It is still possible to use the form in \eqref{eq:density} to express the conditional density function for the spatio-temporal case as
\noindent
\begin{align}
    & f_{u}(t, \boldsymbol{s}) = \lambda_{u}(t, \boldsymbol{s})e^{-\Lambda_{\lambda}(t)},
\end{align}
\noindent
where
\noindent
\begin{align}
    & \Lambda_{\lambda}(t) =  \sum\limits_{u'=1}^U\int_{t_n}^{t}\iint_{\boldsymbol{s'} \in \mathcal{S}}{\lambda_{u'}(t', \boldsymbol{s'})d\boldsymbol{s'}dt'}.
\end{align}
\noindent
To estimate the optimum parameter set $\boldsymbol{\theta}=\{\mathbf{K}^{(\mu)}, \mathbf{K}^{(\gamma)}, \mathbf{W}, \mathbf{\Sigma}^{(\mu)}, \mathbf{\Sigma}^{(\gamma)}\}$, we follow maximum likelihood estimation approach. The negative log-likelihood over the real-life event sequence $\mathcal{E}=\{e_i\}_{i=1}^{N}$ 
is minimized, where $N$ denotes the number of events. The objective is given below:
\noindent
\begin{align}
    \boldsymbol{\hat{\theta}} &= \argmin_{\boldsymbol{\theta}}{\mathcal{L}},
\end{align}
\noindent
where $\mathcal{L}$ is the negative log-likelihood and can be expressed as
\noindent
\begin{align}
{\scriptsize\begin{aligned}
\mathcal{L} &= -\mathrm{log}\left(\prod\limits_{i=1}^{N}f_{u_i}(t_i, \boldsymbol{s_i})\right) =-\sum_{i=1}^{N}\log{\lambda_{u_i}(t_i,\boldsymbol{s_i})}+\sum_{i=1}^{N}\Lambda_{\lambda}(t_i)\end{aligned}}, 
\label{eq:loglik}
\end{align}
\noindent where the second term involving $\Lambda_{\lambda}(t_i)$ can be interpreted as a regularizer, which prevents producing high intensity values over all space defined by $\mathcal{T}$ and $\mathcal{S}$. 

We point out that certain parameters included in $\boldsymbol{\theta}$ are optimized indirectly through reparameterization to handle numerical constraints such as positivity of $\mathbf{K}^{(\cdot)}$ and $\mathbf{W}$, and unique properties of covariance matrices. Methods to handle these constraints during the optimization are explained in Sections~\ref{sec:method_algorithm}.

\subsection{Random Fourier Features}

\label{sec:prob_rff}

Random Fourier features provide an efficient way to approximate the output of a shift-invariant continuous kernel $k(\boldsymbol{x}, \boldsymbol{y})$ with $\boldsymbol{x}, \boldsymbol{y} \in \mathbb{R}^d$~\cite{rahimi_rff}. This technique embeds kernel inputs ($\boldsymbol{x}$ and $\boldsymbol{y}$) into a $D$-dimensional Euclidean inner product space using a transformation matrix $\mathbf{F} \in \mathbb{R}^{d \times D}$ and approximates $k(\boldsymbol{x}, \boldsymbol{y})$ through the inner product of embedded vectors. Although it is widely used to scale up kernel based methods such as SVM for large datasets,~\cite{lu_kernel_scale} we use it to increase the spatial flexibility and replace complex kernel calculations with straightforward matrix multiplications. 

Random Fourier feature-based kernel representation relies on Bochner's Theorem, which states that any bounded, continuous and shift-invariant kernel is a Fourier transform of a bounded non-negative measure~\cite{bochner}. Assuming $p(\cdot)$ is the density function of the spectral measure, the corresponding shift-invariant kernel can be written as
\noindent
\begin{align}
\begin{aligned}
{\small
k(\boldsymbol{x}, \boldsymbol{y}) =  \int_{\mathcal{R}_d}p(\boldsymbol{w})e^{j \boldsymbol{w}^T(\boldsymbol{x}-\boldsymbol{y})}d\boldsymbol{w} = E_{\boldsymbol{w}}[\zeta_{\boldsymbol{w}}(\boldsymbol{x})\zeta_{\boldsymbol{w}}(\boldsymbol{y})^*], \nonumber
}
\end{aligned}
\end{align}
\noindent
where $\zeta_{\boldsymbol{w}}(\boldsymbol{x}) = e^{j \boldsymbol{w}^T\boldsymbol{x}}$, and $c^*$ denotes the complex conjugate of $c \in \mathcal{C}$. Finally, this expression is approximated by its Monte-Carlo estimate as follows,
\noindent
\begin{equation}
\tilde{k}(\boldsymbol{x}, \boldsymbol{y}) = \frac{1}{D}\sum_{i=1}^D z_i (\boldsymbol{x}) z_i (\boldsymbol{y}) = \boldsymbol{z}^T (\boldsymbol{x})\boldsymbol{z}(\boldsymbol{y}) ,\label{eq:krl_appx}
\end{equation}
\noindent
where $z_i(\boldsymbol{x}) = \sqrt{2}\cos{(\boldsymbol{x}^T\boldsymbol{w_i}+b_i)}$ with $\boldsymbol{w_i} \in \mathbb{R}^{d \times 1}$ sampled from $p(\boldsymbol{w})$ and $b_i\sim \text{U}(0, 2\pi)$.

\section{Spatio-Temporal Hawkes Process with Randomized Kernel Representation}

\label{sec:method}

In this section, we describe our method to express the spatial kernels given in \eqref{eq:mu_intro} and \eqref{eq:gamma_intro} with Random Fourier features using  \eqref{eq:krl_appx}, and obtain a neat matrix formulation for the objective function given in \eqref{eq:loglik}. Then, we provide derivative calculations for gradient descent and describe the optimization procedure.

\subsection{Random Fourier Features for Kernel Representation}

\label{sec:method_fourier}

We start with expressing the spatial Gaussian kernel functions of the base and triggering intensity components in \eqref{eq:mu_intro} and \eqref{eq:gamma_intro} using random Fourier features. For given two locations $\boldsymbol{s_i}=[x_i, y_i]^T$ and $\boldsymbol{s_j}=[x_j, y_j]^T$, the result of the Gaussian kernel output in \eqref{eq:g2} can be approximated with the following $D$ dimensional random Fourier approximation~\cite{rahimi_rff} as
\noindent
\begin{equation}\label{eq:rff}
    g_2(\boldsymbol{s_i}, \boldsymbol{s_j}) \approx \frac{1}{2\pi}|\mathbf{\Sigma}|^{-1/2} \boldsymbol{z_i}^T\boldsymbol{z_j},
\end{equation}
where $\boldsymbol{z_i}^T = \sqrt{\frac{2}{D}}\cos{(\boldsymbol{s_i}^T\mathbf{F} + \boldsymbol{b}^T)}$ with $\mathbf{F} \in \mathbb{R}^{2 \times D}$, $\boldsymbol{f_{d}} \sim \mathcal{N}(0, \mathbf{\Tilde{\Sigma}})$ and $b_d$ is sampled uniformly from $[0, 2\pi]$, and $\mathbf{\Tilde{\Sigma}} = \mathbf{\Sigma}^{-1}$.

\begin{figure}[t]
\begin{subfigure}[b]{0.45\columnwidth}
   \centering
\includegraphics[width=\textwidth]{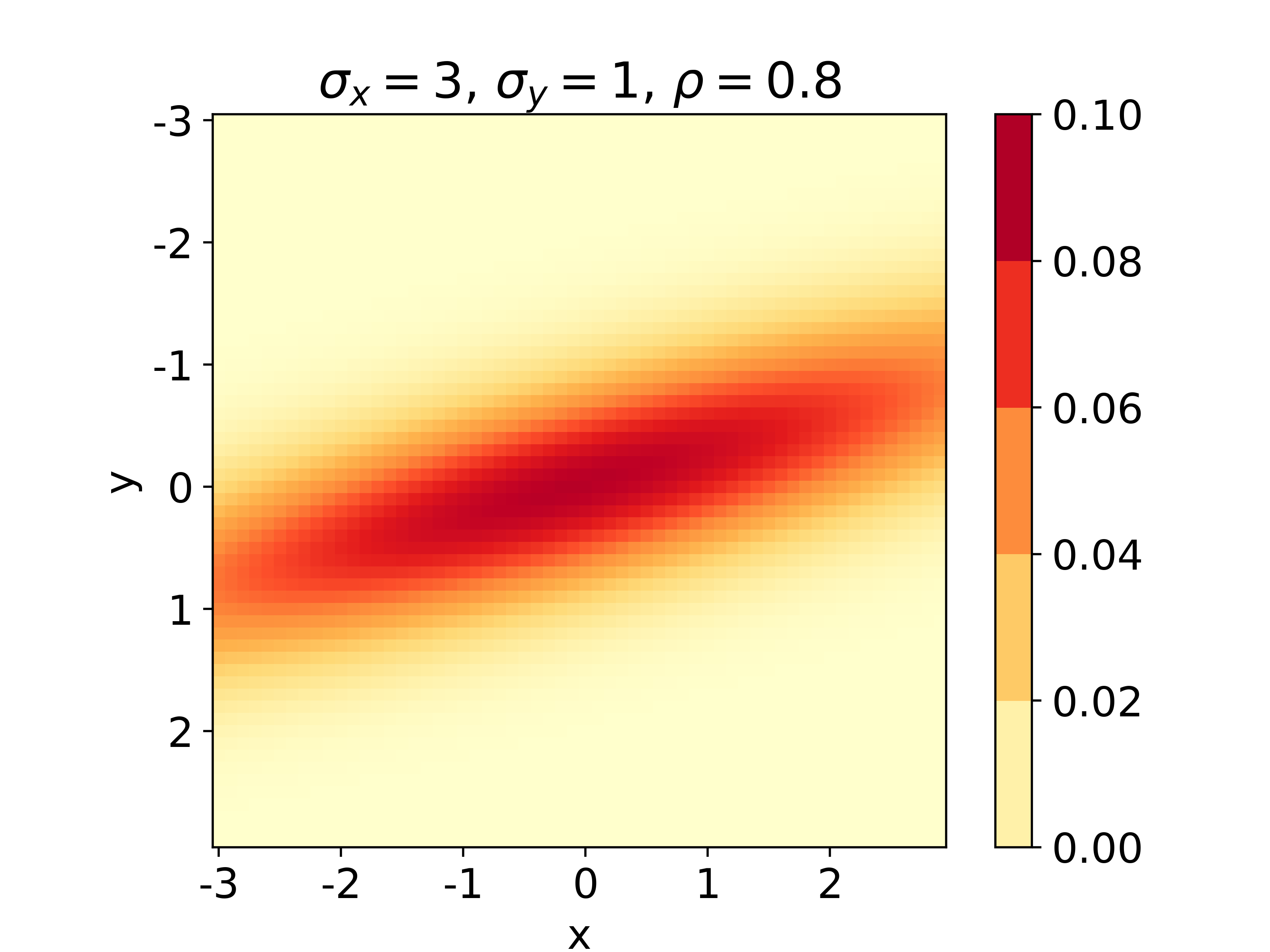}
\caption{} \label{fig:rff1}
\end{subfigure}\hfill
 \begin{subfigure}[b]{0.45\columnwidth}
   \centering
\includegraphics[width=\textwidth]{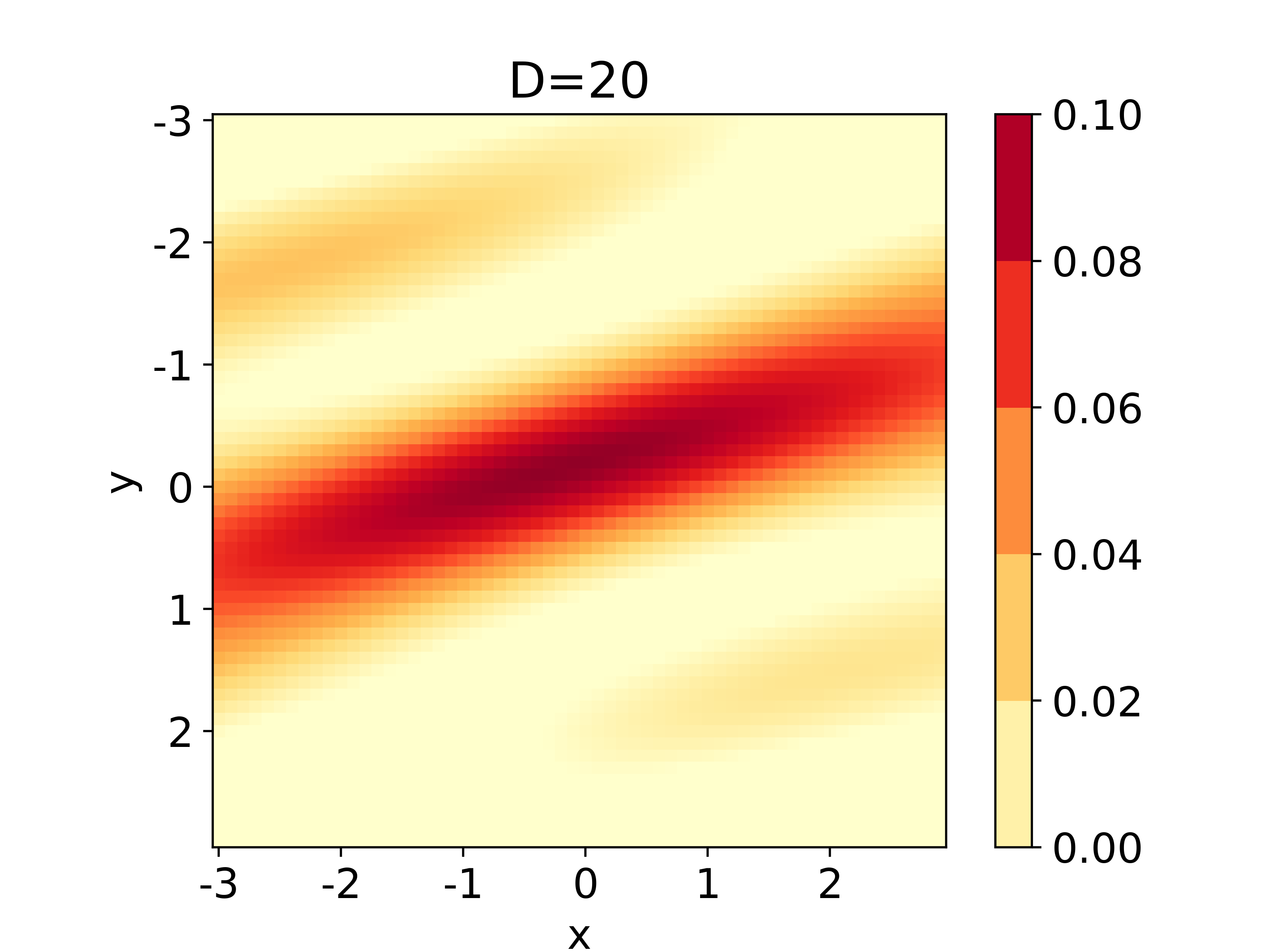}
\caption{} \label{fig:rff2}
\end{subfigure} \\
\begin{subfigure}[b]{0.45\columnwidth}
   \centering
\includegraphics[width=\textwidth]{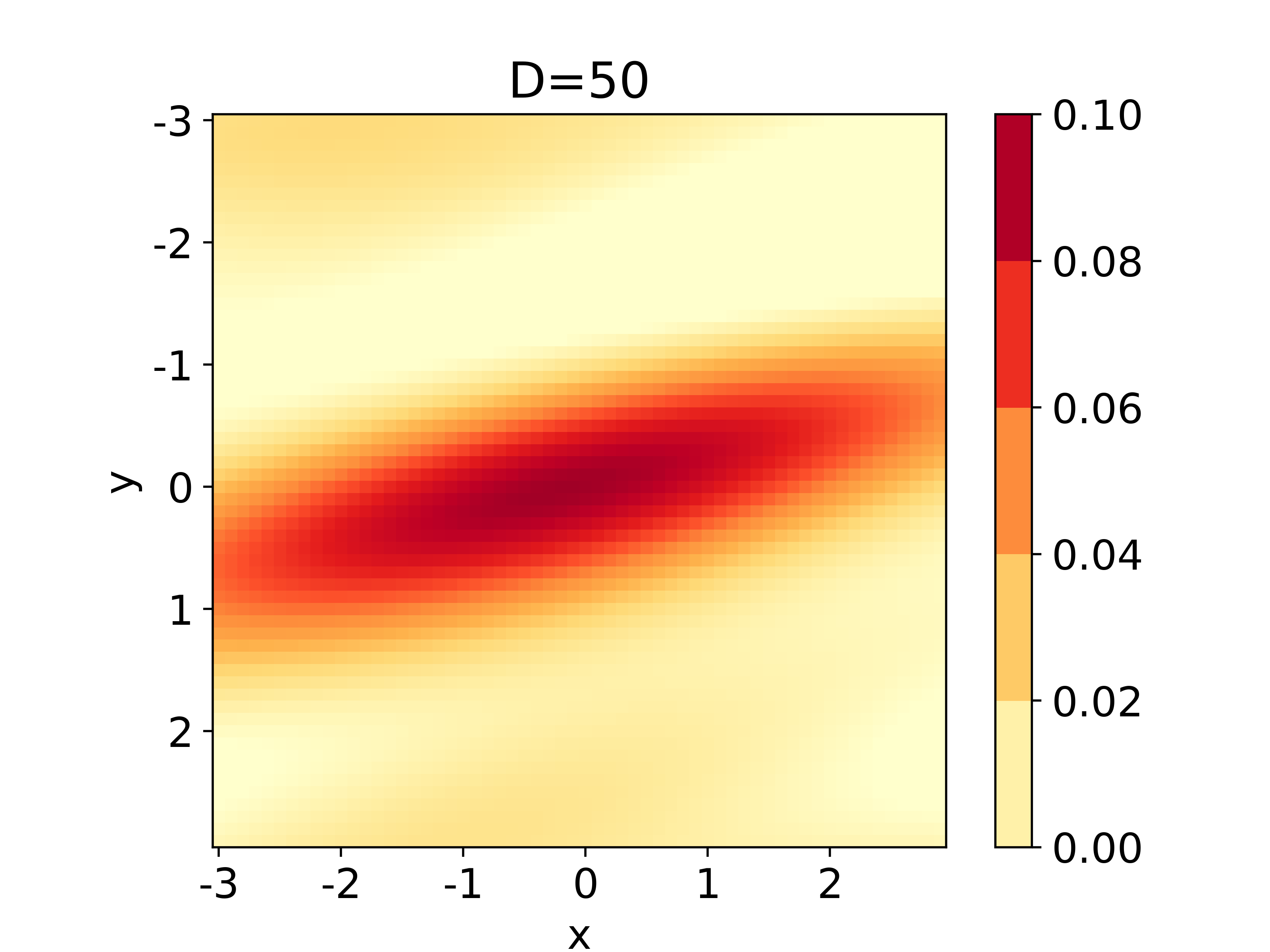}
\caption{} \label{fig:rff3}
\end{subfigure}\hfill
\begin{subfigure}[b]{0.45\columnwidth}
   \centering
\includegraphics[width=\textwidth]{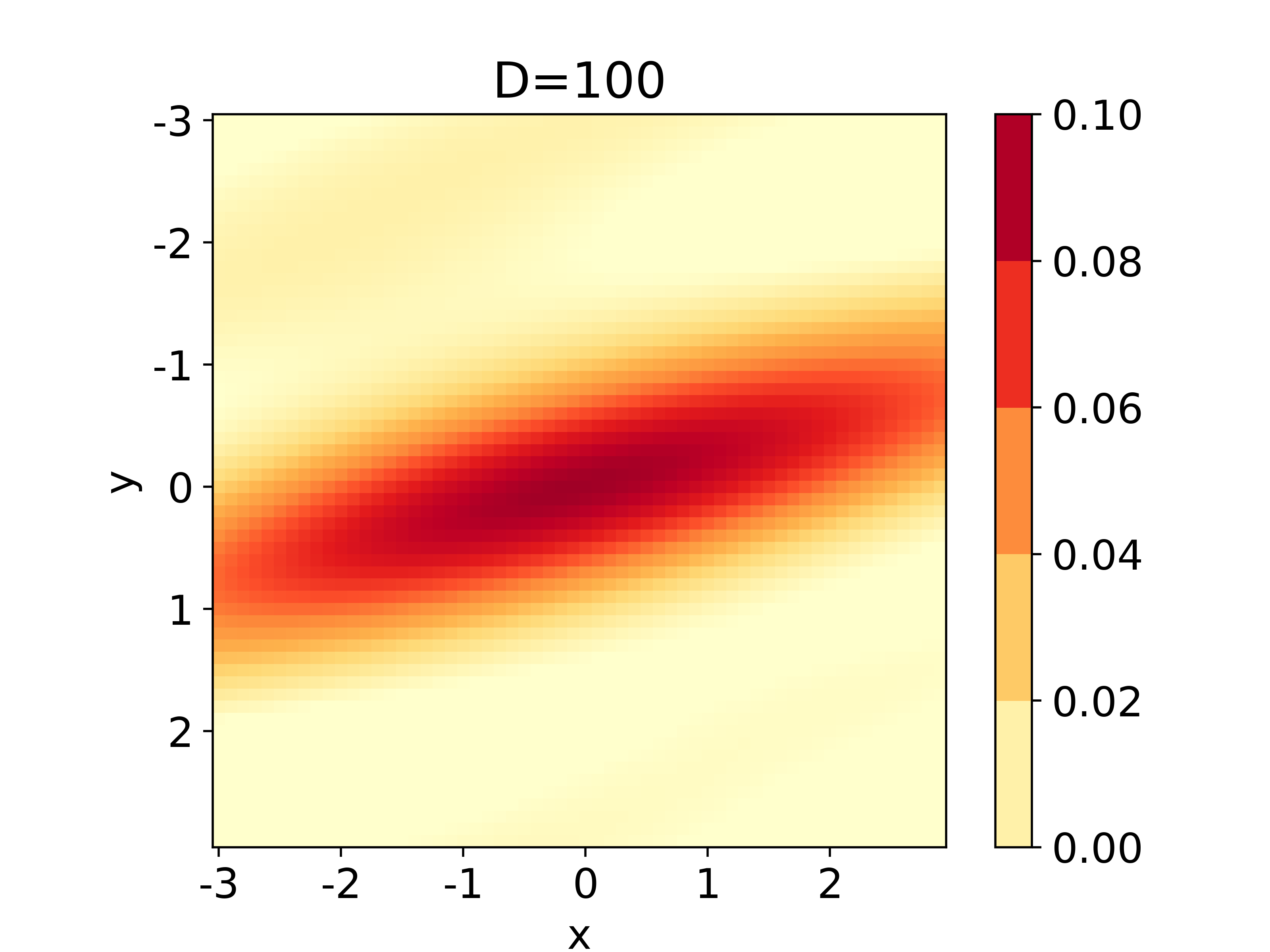}
\caption{} \label{fig:rff4}
\end{subfigure}
\caption{\textbf{(a)} Gaussian kernel with $\sigma_x=3$, $\sigma_y=1$ and $\rho=0.8$, \textbf{(b)}-\textbf{(c)}-\textbf{(d)} Approximated kernels with $20$, $50$ and $100$-dimensional random Fourier features}\label{fig:rff}
\end{figure}

To analyze the behavior of the random Fourier features given in \eqref{eq:rff}, we construct a Gaussian kernel with $\sigma_x=3$, $\sigma_y=1$ and $\rho=0.8$, and perform three approximations with various embedding dimensions ($D = {20, 50, 100}$). We visualize the results in Fig.~\ref{fig:rff}. As the number of dimensions in random Fourier features increases, the approximation becomes more accurate. In the cases when $D$ is small as in Fig.~\ref{fig:rff2}, some randomly repeating artifacts are visible around the kernel.

Since $\mathbf{\Sigma}$ is a positive definite and symmetric matrix as mentioned in the previous section, $\mathbf{\Tilde{\Sigma}}$ is also positive-definite and symmetric. Therefore, we can decompose $\mathbf{\Tilde{\Sigma}}$ using the Cholesky decomposition. and express as $\mathbf{\Tilde{\Sigma}} = \mathbf{\Tilde{C}}\mathbf{\Tilde{C}}^T$, where $\mathbf{\Tilde{C}}$ is a unique, invertible, lower triangular $2\times2$ matrix with real, and positive diagonal entries. Using this decomposition, we obtain the following form for vector embedding:
\noindent
\begin{gather}
    \boldsymbol{z_i}^T = \sqrt{\frac{2}{D}}\cos{(\boldsymbol{s_i}^T\mathbf{\Tilde{C}}\mathbf{U} + \boldsymbol{b}^T)} \label{eq:rff_c},
\end{gather}
where $\mathbf{U} \in \mathbb{R}^{2 \times D}$, and $\boldsymbol{u_{d}} \sim \mathcal{N}(0, \mathbf{I})$. 

We emphasize that $\mathbf{\Tilde{C}}$ introduces certain numerical constraints because of being lower triangular and having positive, real diagonal entries that should be considered during optimization. To handle this issue, we use eigendecomposition of the covariance matrix $\mathbf{\Sigma}$ to express $\mathbf{\Tilde{C}}$ in a simpler form still with constraints but more straightforward to handle:
\noindent
\begin{align}
    {\small
    \mathbf{\Sigma}^{-1} = (\mathbf{V}\mathbf{\Lambda}\mathbf{V}^T)^{-1} = (\mathbf{V}\mathbf{\Lambda}^{-1/2})(\mathbf{\Lambda}^{-1/2}\mathbf{V}^T), 
    }
    \label{eq:eigendecomp}
\end{align}
\noindent
where $\mathbf{V}\mathbf{\Lambda}^{-1/2} = \mathbf{\Tilde{C}}$.

Now, we have two components: the eigenvector matrix $\mathbf{V} \in \mathbb{R}^{2\times2}$ with orthonormality, and the diagonal matrix of eigenvalues $\mathbf{\Lambda}\in \mathbb{R}^{2\times2}$ with positivity constraints. These components can be interpreted as the descriptors of the direction and magnitude of excitation caused by an event. As a result, we obtain the following final form for the vector embedding:
\noindent
\begin{gather}
    \boldsymbol{z_i}^T = \sqrt{\frac{2}{D}}\cos{(\boldsymbol{s_i}^T\mathbf{V}\mathbf{\Lambda^{-1/2}}\mathbf{U} + \boldsymbol{b}^T)}. \label{eq:rff_c}
\end{gather}
\noindent

In addition, we can express $|\mathbf{\Sigma}|^{-1/2}$ in \eqref{eq:rff} in terms of the the diagonal elements of $\mathbf{\Lambda}$ such that $ |\mathbf{\Sigma}|^{-1/2} =\frac{1}{\sqrt{\ell_1 \ell_2}}$, where $\boldsymbol{\ell} = [\ell_1, \ell_2]^T \triangleq [\Lambda_{11}, \Lambda_{22}]^T$ is the vector that consists of the diagonal elements of  $\mathbf{\Lambda}$. Since any term including covariance matrices $\mathbf{\Sigma}^{(\mu)}$ and $\mathbf{\Sigma}^{(\gamma)}$ or their corresponding Cholevsky components can be expressed using $\mathbf{V}^{(\mu)}$, $\mathbf{V}^{(\gamma)}$, $\boldsymbol{\ell}^{(\mu)}$, and $\boldsymbol{\ell}^{(\gamma)}$, we update the notation given for the parameter set as $
    \boldsymbol{\theta} =\{\mathbf{K}^{(\mu)}, \mathbf{K}^{(\gamma)}, \mathbf{W}, \mathbf{V}^{(\mu)}, \mathbf{V}^{(\gamma)}, \boldsymbol{\ell}^{(\mu)}, \boldsymbol{\ell}^{(\gamma)}\}
$. \noindent
\subsection{Matrix Formulations}
\label{sec:method_matrix}
After representing the spatial kernel with the random Fourier feature-based approximation, we formulate the problem in a well-organized matrix form. First, we define the following matrices\footnote{We use $J(t) = \{j | t_j < t\}$ to notate the rows that
belong to the events occurred before t.}:

\noindent
\begin{align}
    & \mathbf{Z}_{J(t)}^{(\cdot)} \triangleq 
    {\begin{bmatrix}
        & \vdots & \\
        \horzbar & \boldsymbol{z}_{j}^{{(\cdot)}^T} & \horzbar \\
        & \vdots &    
    \end{bmatrix}}_{N^{'} \times D} , \label{eq:z} \\
    & \boldsymbol{d}_{J(t)} \triangleq 
    {\begin{bmatrix}
        & \vdots & \\
        & w_{u_j u_i}\exp{(-w_{u_j u_i}(t-t_j))} & \\
        & \vdots &   
    \end{bmatrix}}_{N^{'} \times 1} , \label{eq:d}\\
    & \boldsymbol{Y}_{J(t)} \triangleq 
    {\begin{bmatrix}
        & \vdots & \\
        \horzbar & \boldsymbol{y}_{j}^{T} & \horzbar \\
        & \vdots & 
    \end{bmatrix}}_{N^{'} \times U} , \label{eq:y} \\
    & \mathbf{N}^{(\mu)}(t) \triangleq \frac{1}{2\pi}|\mathbf{\Sigma}|^{-1/2}\mathbf{Z}^{{(\mu)}^T}_{J(t)} \mathbf{Y}_{J(t)}, \label{eq:n_mu} \\
    & \mathbf{N}^{(\gamma)}(t) \triangleq \frac{1}{2\pi}|\mathbf{\Sigma}|^{-1/2}\mathbf{Z}^{{(\gamma)}^T}_{J(t)}\mathrm{diag}(\boldsymbol{d}_{J(t)}) \mathbf{Y}_{J(t)}, \label{eq:n_gamma}
\end{align}
where $\boldsymbol{y}_j^{T}$ is the one-hot vector form of an event type for the $j^{th}$ event. 

Using \eqref{eq:mu_intro}, \eqref{eq:gamma_intro}, \eqref{eq:rff} and \eqref{eq:rff_c}, base and triggering conditional intensity function values for the $i^{th}$ event can be factorized as
\noindent
\begin{align}
    &\mu_{u_i}(\boldsymbol{s_i}) = \frac{1}{T}\boldsymbol{z_i}^{{(\mu)}^T} \mathbf{N}^{(\mu)}(T)\boldsymbol{k}_{u_i}^{(\mu)}, \\
    &\gamma_{u_i}(t_i, \boldsymbol{s_i}) = \boldsymbol{z_i}^{{(\gamma)}^T} \mathbf{N}^{(\gamma)}(t_i)\boldsymbol{k}_{u_i}^{(\gamma)},
\end{align}
where $\boldsymbol{k}_{u_i}^{(\cdot)}$ is the $u_i^{th}$ column of $\mathbf{K}^{(\cdot)}$, which contains the effects of other event types over the event type of the $i^{th}$ event. Finally, using \eqref{eq:n_mu} and \eqref{eq:n_gamma}, the conditional intensity values for $\mathcal{E}$ given in \eqref{eq:loglik} can be expressed in the following matrix form:
\begin{equation}  \label{eq:A_mat}
    \mathbf{A} \triangleq
    \begin{bmatrix}
        & \vdots & \\
        \horzbar & \boldsymbol{\lambda}(t_i, \boldsymbol{s_i}) & \horzbar \\
        & \vdots &    
    \end{bmatrix} = \mathbf{Q}^{(\mu)}\mathbf{K}^{(\mu)} + \mathbf{Q}^{(\gamma)}\mathbf{K}^{(\gamma)},
\end{equation}
\noindent
where
\noindent
\begin{align}
    \mathbf{Q}^{(\mu)} \triangleq 
    \begin{bmatrix}
        & \vdots & \\
        \horzbar & \frac{1}{T}\boldsymbol{z_i}^{{(\mu)}^T}\mathbf{N}^{(\mu)}(T) & \horzbar \\
        & \vdots &         
    \end{bmatrix}_{N\times U} \nonumber
\end{align}
contains the relation between the $i^{th}$ event and other events for base intensity, and
\noindent
\begin{align}
    \mathbf{Q}^{(\gamma)} \triangleq 
    \begin{bmatrix}
        & \vdots & \\
        \horzbar & \boldsymbol{z_i}^{{(\gamma)}^T}\mathbf{N}^{(\gamma)}(t_i) & \horzbar \\
        & \vdots &         
    \end{bmatrix}_{N\times U} \nonumber
\end{align}
contains the relation between the $i^{th}$ event and past events for triggering intensity at each row.

Once obtaining the matrix-form expression for the conditional intensity in \eqref{eq:A_mat}, we analytically derive the integral output to obtain the closed-form expression for the second term in \eqref{eq:loglik} as
\noindent
\begin{align}
    \Lambda_{\lambda}(t_i) &=  \sum\limits_{u'=1}^U\int\displaylimits_{t_{i-1}}^{t_i}\iint\displaylimits_{\boldsymbol{s'} \in \mathcal{S}}{\lambda_{u'}(t', \boldsymbol{s'})d\boldsymbol{s'}dt'} \\
    & \approx \sum\limits_{u'=1}^U\int\displaylimits_{t_{i-1}}^{t_i}\iint\displaylimits_{\boldsymbol{s'} \in \mathbb{R}^2}{\mu_{u'}(\boldsymbol{s'})d\boldsymbol{s'}dt'} \nonumber \\ 
    &\quad + \sum\limits_{u'=1}^U\int\displaylimits_{t_{i-1}}^{t_i}\iint\displaylimits_{\boldsymbol{s'} \in \mathbb{R}^2}{\gamma_{u'}(t', \boldsymbol{s'})d\boldsymbol{s'}dt'} \nonumber \\
    & \approx \sum\limits_{u'=1}^U\int\displaylimits_{t_{i-1}}^{t_i}\iint\displaylimits_{\boldsymbol{s'} \in \mathbb{R}^2}{\frac{1}{T}\sum_{j=1}^N k_{u_ju'}^{(\mu)} g_2^{(\mu)}(\boldsymbol{s'}, \boldsymbol{s_j})d\boldsymbol{s'}dt'} \nonumber \\ 
    & \quad + \sum\limits_{u'=1}^U\int\displaylimits_{t_{i-1}}^{t_i}\iint\displaylimits_{\boldsymbol{s'} \in \mathbb{R}^2} \sum_{j|t_j<t'} \Big( k_{u_ju'}^{(\gamma)}g_1(t', t_j, u', u_j) \nonumber \\ 
    & \quad \times g_2^{(\gamma)}(\boldsymbol{s'}, \boldsymbol{s_j}) \Big) d\boldsymbol{s'}dt' \nonumber \\
    & \approx \frac{t_i-t_{i-1}}{T}\sum_{j=1}^N\sum_{u'=1}^Uk_{u_j u'}^{(\mu)} \nonumber \\ 
    & \quad +\sum_{j|t_j<t_i}\sum_{u'=1}^Uk_{u_j u'}^{(\gamma)}(e^{-w_{u_j u'}(t_{i-1}-t_j)}-e^{-w_{u_j u'}(t_i-t_j)}) \nonumber,
\end{align}
\noindent
where we approximate $\mathcal{S}$ with $\mathbb{R}^2$ since the boundary effects will have a negligible effect over the integral value. Then, the summation of $\Lambda_{\lambda}(t_i)$ for consecutive events is expressed as
\begin{align}
    \sum_{i=n-k}^n \Lambda(t_{i}) &= \frac{t_n-t_{n-k-1}}{T}\sum_{j=1}^N\sum_{u'=1}^Uk_{u_j u'}^{(\mu)} \nonumber \\ 
    &- \sum_{j|t_j<t_n}\sum_{u'=1}^Uk_{u_j u'}^{(\gamma)}(e^{-w_{u_j u'}(t_n-t_j)}) \nonumber \\
    &+ \sum_{j|t_j<t_{n-k-1}}\sum_{u'=1}^Uk_{u_j u'}^{(\gamma)}(e^{-w_{u_j u'}(t_{n-k-1}-t_j)}) \nonumber \\
    &+ \sum_{j|t_{n-k-1}\leq t_j<t_n}\sum_{u'=1}^Uk_{u_j u'}^{(\gamma)} \label{eq:sum_consec}
\end{align}
\noindent
for $0\leq k < i$ and $t_0=0$. Here, we utilize the relation between consecutive terms, which cancels out most of the intermediate outputs. Inserting $n=N$ and $k=N-1$ into \eqref{eq:sum_consec} yields the following:
\begin{align} \label{eq:reg_term}
{\scriptsize
    \begin{aligned}
    R \triangleq \sum_{i=1}^N \Lambda(t_{i}) = \sum_{j=1}^N\sum_{u'=1}^Uk_{u_j u'}^{(\mu)} + k_{u_j u'}^{(\gamma)}(1-e^{-w_{u_j u'}(T-t_j)}), 
    \end{aligned}}
\end{align}
where $R$ is defined as the second term in \eqref{eq:loglik}, and has a suppressing effect over excitation matrices. 

Finally, using \eqref{eq:loglik}, \eqref{eq:A_mat}, and \eqref{eq:reg_term}, we can express the negative log-likelihood as
\noindent
\begin{equation} \label{eq:opt}
    \mathcal{L} = -\text{sum}(\log{(\mathbf{A})}\odot \mathbf{Y}) + R.
\end{equation}

In order to minimize the negative log-likelihood expressed in \eqref{eq:opt}, we employ gradient descent through the back propagation of derivatives $\frac{\partial\mathcal{L}}{\partial\boldsymbol{\theta}} = \{ \frac{\partial\mathcal{L}}{\partial\mathbf{K}^{(\mu)}}, \frac{\partial\mathcal{L}}{\partial\mathbf{K}^{(\gamma)}}, \frac{\partial\mathcal{L}}{\partial\mathbf{W}}, \frac{\partial\mathcal{L}}{\partial\mathbf{V}^{(\mu)}}, \frac{\partial\mathcal{L}}{\partial\mathbf{V}^{(\gamma)}},
\frac{\partial\mathcal{L}}{\partial\boldsymbol{l}^{(\mu)}},
\frac{\partial\mathcal{L}}{\partial\boldsymbol{l}^{(\gamma)}}\}$. We provide the equations for these gradients in Appendix~\ref{sec:app_a}.

\subsection{Optimization Algorithm}
\label{sec:method_algorithm}

Here, we detail the optimization procedure to minimize the negative log-likelihood $\mathcal{L}$ expressed in \eqref{eq:opt}. We adapt mini-batch gradient descent into our problem with a slightly modified batch generation procedure as explained in Algorithm \ref{alg:mbgd}. We also follow a training procedure with early stopping that stops the iterations if the model does not improve during $k$  consecutive steps in terms of negative log-likelihood.

\begin{algorithm}
{\small
\caption{Mini-Batch Gradient Descent with Random Fourier Features (RFF-GD).}
\label{alg:mbgd}
\begin{algorithmic}
\REQUIRE: $\boldsymbol{\theta}$ (Initial parameter set), $\mathcal{E}_{train}$ (Event sequence for training), $\mathcal{E}_{val}$ (Event sequence for validation), $b$ (batch size), $\eta$ (learning rate), max\_epoch (number of maximum epochs) and $\pi=False$ (early stopping flag)
\WHILE{$\texttt{epoch} < \text{max\_epoch}$}
\STATE $\texttt{step} \leftarrow 0$
\WHILE{$\texttt{step} < N_{train}/b$}
\STATE Sample $i_s$ uniformly from $\{1, 2, \hdots N_{train}-b\}$.
\STATE $i_e \leftarrow i_s + b$
\STATE $X \leftarrow \{e_{{train}_i}\}_{i=i_s}^{i_e}$
\FORALL{$\theta_k \in \boldsymbol{\theta}$}
\STATE Update $\theta_k$ using \eqref{eq:deriv_start}-\eqref{eq:deriv_end}.
\ENDFOR
\STATE $\texttt{step} \leftarrow \texttt{step} + 1$
\ENDWHILE
\STATE Calculate $\mathcal{L}$ over $\mathcal{E}_{val}$ with \eqref{eq:loglik}, and update $\pi$ based on early stopping criteria.
\IF{$\pi$}
\RETURN $\boldsymbol{\theta}$
\ENDIF
\STATE $\texttt{epoch} \leftarrow \texttt{epoch} + 1$
\ENDWHILE
\RETURN $\boldsymbol{\theta}$
\end{algorithmic}
}
\end{algorithm}

As mentioned before, certain parameters in $\boldsymbol{\theta}$ have constraints. The elements of the excitation matrices $\mathbf{K}^{(\mu)}$ and $\mathbf{K}^{(\gamma)}$, the decay matrix $\mathbf{W}$, and the eigenvalue vectors of covariance matrices, $\boldsymbol{\ell}^{(\mu)}$ and $\boldsymbol{\ell}^{(\gamma)}$ have to be positive. To satisfy these conditions, we simply introduce the following intermediate variables and perform gradient descent over the unconstrained parameters $\mathbf{\Tilde{K}}^{(\cdot)}$, $\mathbf{\Tilde{W}}$ and $\mathbf{\boldsymbol{\Tilde{\ell}}^{(.)}}$:
\noindent
\begin{align}
    & \mathbf{K}^{(\cdot)} = \phi(\mathbf{\Tilde{K}}^{(\cdot)}) = \frac{1}{s}\log{(1+e^{{s\mathbf{\Tilde{K}}^{(\cdot)}}})}, \nonumber \\
    & \mathbf{W} = \phi(\mathbf{\Tilde{W}}) = \frac{1}{s}\log{(1+e^{{s\mathbf{\Tilde{W}}}})}, \nonumber \\
    & \mathbf{\boldsymbol{\ell}^{(.)}} = \phi(\boldsymbol{\Tilde{\ell}}^{(.)}) = \frac{1}{s}\log{(1+e^{{s\boldsymbol{\Tilde{\ell}}^{(.)}}})}, \nonumber 
\end{align}
\noindent
where $\phi$ is the soft-plus function parametrized by $s$. Soft-plus function provides a differentiable and smooth approximation of rectified linear unit function ($\sigma_{\text{ReLU}}(x)=\text{max}(0, x)$) such that as $s \rightarrow \infty$, $\phi \rightarrow \sigma_{\text{ReLU}}$~\cite{softplus}.

Other constrained parameters are the eigenvector matrices $\mathbf{V}^{(\mu)}$ and $\mathbf{V}^{(\gamma)}$, which have to be orthonormal due to the eigendecomposition in \eqref{eq:eigendecomp}. We employ projected gradient descent to meet this limitation by using the following update rule:
\noindent
\begin{align}
    \mathbf{V}^{(.)}_{t+1} = \Pi_{\chi}\left(\mathbf{V}^{(.)}_{t} - \eta\frac{\partial \mathcal{L}}{\partial \mathbf{V}^{(.)}_{t}}\right), \; \forall t \geq 1 , \nonumber
\end{align}
\noindent
where $\chi = \{\mathbf{X} \; | \; \mathbf{X} \in \mathbb{R}^2, \, \mathbf{X}^T\mathbf{X} = \mathbf{I}\}$ is the convex set of orthonormal matrices. Here, $\Pi_{\chi}$ projects the updated parameter to $\chi$ through solving the following minimization problem known as orthonogal Procrustes problem~\cite{procrustes}:
\noindent
\begin{align}
	\Pi_{\chi}(\mathbf{\Tilde{X}}) &= \argmin_{\mathbf{X}}(||\mathbf{X} - \mathbf{\Tilde{X}}||_F) \;\; \text{subject to} \;\; \mathbf{X} \in \chi \nonumber\\ 
	&= \mathbf{U}\mathbf{V}^T 	\label{eq:deriv_end},
\end{align}
\noindent
where $||\cdot||_F$ denotes the Frobenius norm, and $\mathbf{\Tilde{X}} = \mathbf{U}\mathbf{\Sigma}\mathbf{V}^T$.

\section{Experiments}
\label{sec:experiments}

In this section, we report the results of our method in terms of fitting performance on synthetic and real-life datasets. We generate three synthetic datasets to analyze the behavior of the proposed approach in a controlled manner. Then, we demonstrate the performance of our method in two real-life datasets and compare it with the EM algorithm~\cite{yuan_mvhawkes} and stochastic declustering~\cite{zhuang_stodecl}. We also analyze the effect of the randomized feature space size on our performance. Finally, we perform event analysis through the interpretation of the inferred process parameters.

\subsection{Synthetic Dataset Experiments}

We first introduce a thinning-based algorithm to simulate synthetic event sequences according to given process parameters. Then, we evaluate two simple baseline approaches in addition to our method over three different simulations.

\subsubsection{Spatio-Temporal Thinning Algorithm for Simulations}

In order to simulate a spatio-temporal Hawkes process, we use the thinning algorithm~\cite{lewis_pp}, which applies rejection sampling over pre-sampled points. Unlike the extension of the thinning algorithm for spatio-temporal case in~\cite{rein_stpp}, we have multiple event types. The details are given in Algorithm \ref{alg:sim}. 

To apply rejection sampling, we need an upper bound for the conditional intensity function,
\begin{align}
{\small
    \Bar{\lambda} \triangleq \max\Bigg(\sum_{u=1}^U\lambda_u(t', \boldsymbol{s'})\Bigg) \; \text{for} \; t' \in [t, +\infty) \; \text{and} \; \boldsymbol{s'} \in \mathcal{S}} \nonumber
\end{align}
\noindent
such that $\lambda(t', \boldsymbol{s'}) < \Bar{\lambda}$ for all $t\geq t'$ and $\boldsymbol{s'} \in \mathcal{S}$. Since the conditional intensity decreases in time exponentially, upper bound will take place at time $t$, so we can express $\Bar{\lambda}$ as
\noindent
\begin{align}
    {\small
    \begin{aligned}
    \Bar{\lambda} = \sum_{u=1}^U \max(\mu_u(\boldsymbol{s'}) + \sum_{j|t_j<t}k_{u_ju}^{(\gamma)}g_1(t, t_j,u,u_j)g_2^{(\gamma)}(\boldsymbol{s'}, \boldsymbol{s_j}))
    \end{aligned}}
    \label{eq:sim_approx}
\end{align}
\noindent
for $\boldsymbol{s'} \in \mathcal{S}$. We perform calculations for densely sampled spatial points over $\mathcal{S}$ at time $t$, and take the maximum value due to the non-monotonous structure of the conditional intensity function over the spatial domain. Moreover, older events will have significantly less effect over the total sum in \eqref{eq:sim_approx} due to the exponential decay kernel. Thus, to make the simulation process computationally efficient, we ignore the triggering effects of the events that occurred before a particular temporal offset ($\tau=100$). We observe no difference in the simulation with this modification and the algorithm is robust to the selection of this parameter.

\begin{algorithm}
{\small
\caption{Thinning Algorithm for Spatio-Temporal Hawkes Process Simulation}
\label{alg:sim}
\begin{algorithmic}
\REQUIRE: $\lambda$ (Conditional intensity function), $\mathcal{T}$ (Temporal space) and $\mathcal{S}$ (Spatial space), $t=0$, $i=1$, $\mathcal{E}=\{\}$
\WHILE{True}
\STATE Estimate $\Bar{\lambda} = \max(\sum_{u=1}^U\lambda_u(t', \boldsymbol{s'}))$ for $t' \in [t, +\infty)$ and $\boldsymbol{s'} \in \mathcal{S}$ by \eqref{eq:sim_approx}.
\STATE Draw $q \sim \mathcal{U}(0, 1)$
\STATE $\Delta t \Leftarrow - \log(q) / \Bar{\lambda}$
\STATE $t \Leftarrow t + \Delta t$
\IF{$t > T$}
\RETURN $\mathcal{E}$
\ENDIF
\STATE Draw $\boldsymbol{s} \sim \mathcal{U}(\mathcal{S})$, $v \sim \mathcal{U}(0, 1)$.
\STATE Calculate $\lambda(t, \boldsymbol{s}) = \sum_{u=1}^U\lambda_u(t, \boldsymbol{s})$.
\IF{$\lambda(t, \boldsymbol{s}) > v\Bar{\lambda}$}
\STATE Draw $u \sim p(u)$, where $p(u)=\frac{e^{-\lambda_u}}{\sum_{u'=1}^U e^{-\lambda_{}u'}}$
\STATE $e_i \Leftarrow [t, \boldsymbol{s}, u]$
\STATE $\mathcal{E} \Leftarrow \mathcal{E} \cup e_i$
\STATE $i \Leftarrow i + 1$
\ENDIF
\ENDWHILE
\end{algorithmic}
}
\end{algorithm}

To generate the type of the thinned event in Algorithm \ref{alg:sim}, we apply the thinning procedure over the total conditional intensity and then draw the event type stochastically from the generated $p(u)$ for the generated spatio-temporal point. Instead of running rejection sampling for each event type separately, this procedure provides an efficient and convenient way to generate spatio-temporal Hawkes process with multiple event types.

\renewcommand{\tabcolsep}{2pt}
\begin{table}[h]
\centering
\captionsetup{justification=centering}
\caption{\textsc{Simulation Configurations. $\mathbf{K}^{(\mu)}$ and $\mathbf{K}^{(\gamma)}$ are the Excitation Matrices for the Base and Triggering Intensities, $\mathbf{\Sigma}^{(\mu)}$ and $\mathbf{\Sigma}^{(\gamma)}$ are the Covariance Matrices of the Spatial Kernels for the Base and Triggering Intensities, and $\mathbf{W}$ is the Weight Decay Matrix of the Temporal Kernel.}}
\label{tab:sims}
\begin{adjustbox}{width=\columnwidth,center}
\renewcommand{\arraystretch}{1.2}
 \begin{tabular}{|c||c|c|c|c|c|} 
  \hline
  ID & $\mathbf{K}^{(\mu)}$ & $\mathbf{K}^{(\gamma)}$ & $\mathbf{W}$ & $\mathbf{\Sigma}^{(\mu)}$ & $\mathbf{\Sigma}^{(\gamma)}$\\
  \hline
  1 & $\big[ \begin{array}{c} 0.01 \end{array} \big]$ & $\left[ \begin{array}{c} 0 \end{array} \right]$ & - & $\Big[ \begin{array}{cc} 10 & 0 \\ 0 & 10 \end{array} \Big]$ & -\\
  \hline
  2 & $\Big[ \begin{array}{cc} 0.01 & 0.005 \\ 0.01 & 0.02 \end{array} \Big]$ & $\Big[ \begin{array}{cc} 0 & 0 \\ 0 & 0 \end{array} \Big]$ & - & $\Big[ \begin{array}{cc} 0.01 & 0 \\ 0 & 0.01 \end{array} \Big]$ & -\\
  \hline
  3 & $\Big[ \begin{array}{cc} 0.01 & 0.005 \\ 0.01 & 0.02 \end{array} \Big]$ & $\Big[ \begin{array}{cc} 1 & 0.5 \\ 1 & 2 \end{array} \Big]$ & $\Big[ \begin{array}{cc} 2 & 1 \\ 1 & 4 \end{array} \Big]$ & $\Big[ \begin{array}{cc} 0.01 & 0 \\ 0 & 0.01 \end{array} \Big]$ & $\Big[ \begin{array}{cc} 0.04 & 0 \\ 0 & 0.04 \end{array} \Big]$\\
  \hline
\end{tabular}
\end{adjustbox}
\end{table}

We simulate realizations with $T=100000$ and $\mathcal{S} = [[-1, 1], [-1, 1]]$. Each simulation is set to different parameters to analyze the behavior in different cases. Table~\ref{tab:sims} shows the parameter sets used for simulations.

\subsubsection{Synthetic Dataset Performance}
We also evaluate two baseline processes with more basic forms compared to the spatio-temporal Hawkes process to analyze the behavior of the proposed framework under different scenarios. First, we consider the Poisson process, where each event type has a constant intensity ($\lambda_u$) over the spatio-temporal space:
\noindent
\begin{align} \label{eq:base1}
    \lambda_u(t, \boldsymbol{s}) = \mu_u,
\end{align} 
\noindent
where $\mu_u$ is the base intensity for the $u^{th}$ event type.

Second, we allow the conditional intensity to be locally variant, but temporally constant by setting $\mathbf{K}^{(\gamma)}$ to be a zero matrix. This can be interpreted as a spatially inhomogeneous Poisson process. For this baseline, the conditional intensity has the following form:
\noindent
\begin{align} \label{eq:base2}
    \lambda_u(t, \boldsymbol{s}) = & \mu_{u}(\boldsymbol{s}) = \frac{1}{T}\sum_{j=1}^N k_{u_ju}^{(\mu)} g_2^{(\mu)}(\boldsymbol{s}, \boldsymbol{s_j}).
\end{align} 
\noindent

For all experiments, we divide each event sequence into training ($80\%$) and test ($20\%$) sets. We use $10\%$ of the training set for the hyperparameter search and early stopping. We obtain maximum likelihood estimates of the process parameters using Algorithm \ref{alg:mbgd} and report the negative log-likelihoods on the training and test sets. We also investigate the Akaike's Information Criterion (AIC)~\cite{akaike}, which is also shown to be a consistent measure while evaluating point process models and preferred in numerous studies in the point process literature including~\cite{Ogata1988,hawkes_ic,Ogata1998}. AIC is defined as follows~\cite{akaike}:
\begin{align}
    &\text{AIC} = 2\mathcal{L} + 2k
\end{align}
where $\mathcal{L}$ is the negative log-likelihood and $k$ is the number of parameters. Although these criteria are maximum likelihood driven and tend to choose the model which fits to the data best, they also penalize the number of parameters to address complexity. Since spatio-temporal Hawkes modeling have more parameters as provided in Table \ref{tab:sim_res}, AIC penalizes it more heavily. Both measures indicate better performance at lower values.
\renewcommand{\tabcolsep}{1.5pt}
\begin{table}[h]
\centering
\captionsetup{justification=centering}
\caption{\textsc{Training performance of Our Algorithm with Poisson \eqref{eq:base1}, Spatial Poisson \eqref{eq:base2} and Spatio-Temporal Hawkes \eqref{eq:lambda_generic} Process Modeling on Synthetic Datasets. Synthetic Data are Simulated with the Parameter Configurations given in Table~\ref{tab:sims} (\smaller{$p$: Number of parameters, $\overline{\mathcal{L}}$}: Negative log-likelihood per event).}}
\begin{adjustbox}{max width=\columnwidth}
\begin{tabular}{|c|c|c|c|c|c|c|c|c|c|}
\hline
\multirow{3}{*}{Model} & \multicolumn{9}{c|}{Simulation ID} \\ \cline{2-10} 
 & \multicolumn{3}{c|}{1} & \multicolumn{3}{c|}{2} & \multicolumn{3}{c|}{3} \\ \cline{2-10} 
\rule{0pt}{8pt} & $p$ & $\overline{\mathcal{L}}$ & AIC & $p$ & $\overline{\mathcal{L}}$ & AIC & $p$ & $\overline{\mathcal{L}}$ & AIC \\ \hline
Poisson & $1$ & $-1.51$ & $-800.94$ & $2$ & $-0.04$ & $-175.36$ & $2$ & $1.39$ & $4927.38$ \\ \hline
Spatial Poisson & $7$ & $-1.44$ & $-1705.36$ & $10$ & $-0.77$ & $-3432.68$ & $10$ & $0.13$ & $480.46$ \\ \hline
ST-Hawkes & $15$ & $-1.57$ & $-1844.58$ & $24$ & $-0.78$ & $-3449.52$ & $24$ & $-2.39$ & $-8417.38$ \\ \hline
\end{tabular}
\end{adjustbox}\label{tab:sim_res}
\end{table}
\noindent
The results for all model-simulation pairs are shown in Table~\ref{tab:sim_res}. All experiments are repeated $10$ times. We highlight that synthetic event sequences are scaled temporally before training to prevent numerical instability issues related to very low/high temporal space size, and report comparable results among all simulations. We shrink the event times such that the average temporal distance between consecutive events becomes 1 unit. We also normalize the resulting negative log-likelihood by dividing it by the number of events, and consider the negative log-likelihood per event to provide comparability across datasets. 
\noindent
\renewcommand{\tabcolsep}{6pt}
\begin{table}[h]
\centering
\captionsetup{justification=centering}
\caption{\textsc{Test performance of Our Algorithm with Poisson \eqref{eq:base1}, Spatial Poisson \eqref{eq:base2} and Spatio-Temporal Hawkes \eqref{eq:lambda_generic} Process Modeling on Synthetic Datasets in terms of Negative Log-likelihood per Event.}}
\begin{tabular}{|c|c|c|c|}
\hline
\multirow{2}{*}{Model} & \multicolumn{3}{c|}{Simulation ID} \\ \cline{2-4} 
 & 1 & 2 & 3 \\ \hline
Poisson & $-1.43$ & $0.09$ & $1.38$ \\ \hline
Spatial Poisson & $-1.51$ & $-0.59$ & $0.32$ \\ \hline
ST-Hawkes & $-1.59$ & $-0.61$ & $-2.18$ \\ \hline
\end{tabular}\label{tab:sim_res_test}
\end{table}
\noindent
As seen in Table~\ref{tab:sim_res}, all models perform similarly in the first simulation since the conditional intensity function is spatio-temporally homogeneous. In the second simulation, the simple Poisson process performs worse because the spatial triggering effect is not included in its modeling. In the third simulation, due to the introduced temporal excitation, spatio-temporal (ST) Hawkes performs significantly better than others thanks to its capability to express spatial and temporal inhomogeneity in the conditional intensity function. Hence, we conclude that our algorithm performs consistent with different modeling choices. In Table \ref{tab:sim_res_test}, we provide the negative log-likelihood per event values obtained on the test set. The results demonstrate the generalization capability of our method since there is no considerable gap between training and test performances. We also provide the recovered intensity function parameters for the synthetic data experiment, which simulates a spatio-temporal Hawkes process, in Table \ref{tab:recovery}.

\renewcommand{\tabcolsep}{1.5pt}
\begin{table}[h]
\centering
\captionsetup{justification=centering}
\caption{\textsc{Estimated Parameters for the 3rd Simulation}}
\label{tab:recovery}
\begin{adjustbox}{width=\columnwidth,center}
\renewcommand{\arraystretch}{1.2}
 \begin{tabular}{|c|c|c|c|c|} 
  \hline
  $\mathbf{K}^{(\mu)}$ & $\mathbf{K}^{(\gamma)}$ & $\mathbf{W}$ & $\mathbf{\Sigma}^{(\mu)}$ & $\mathbf{\Sigma}^{(\gamma)}$\\
  \hline
  $\Big[ \begin{array}{cc} 0.018 & 0.003 \\ 0.011 & 0.038 \end{array} \Big]$ & $\Big[ \begin{array}{cc} 1.111 & 0.44 \\ 0.983 & 1.976 \end{array} \Big]$ & $\Big[ \begin{array}{cc} 1.78 & 0.943 \\ 0.9 & 3.96 \end{array} \Big]$ & $\Big[ \begin{array}{cc} 0.008 & 0.003 \\ 0.003 & 0.011 \end{array} \Big]$ & $\Big[ \begin{array}{cc} 0.039 & 0.001 \\ 0.001 & 0.038 \end{array} \Big]$\\
  \hline
\end{tabular}
\end{adjustbox}
\end{table}

\begin{figure*}[t]
\centering
\begin{subfigure}[t]{0.45\textwidth}
   \centering
\includegraphics[width=0.9\textwidth]{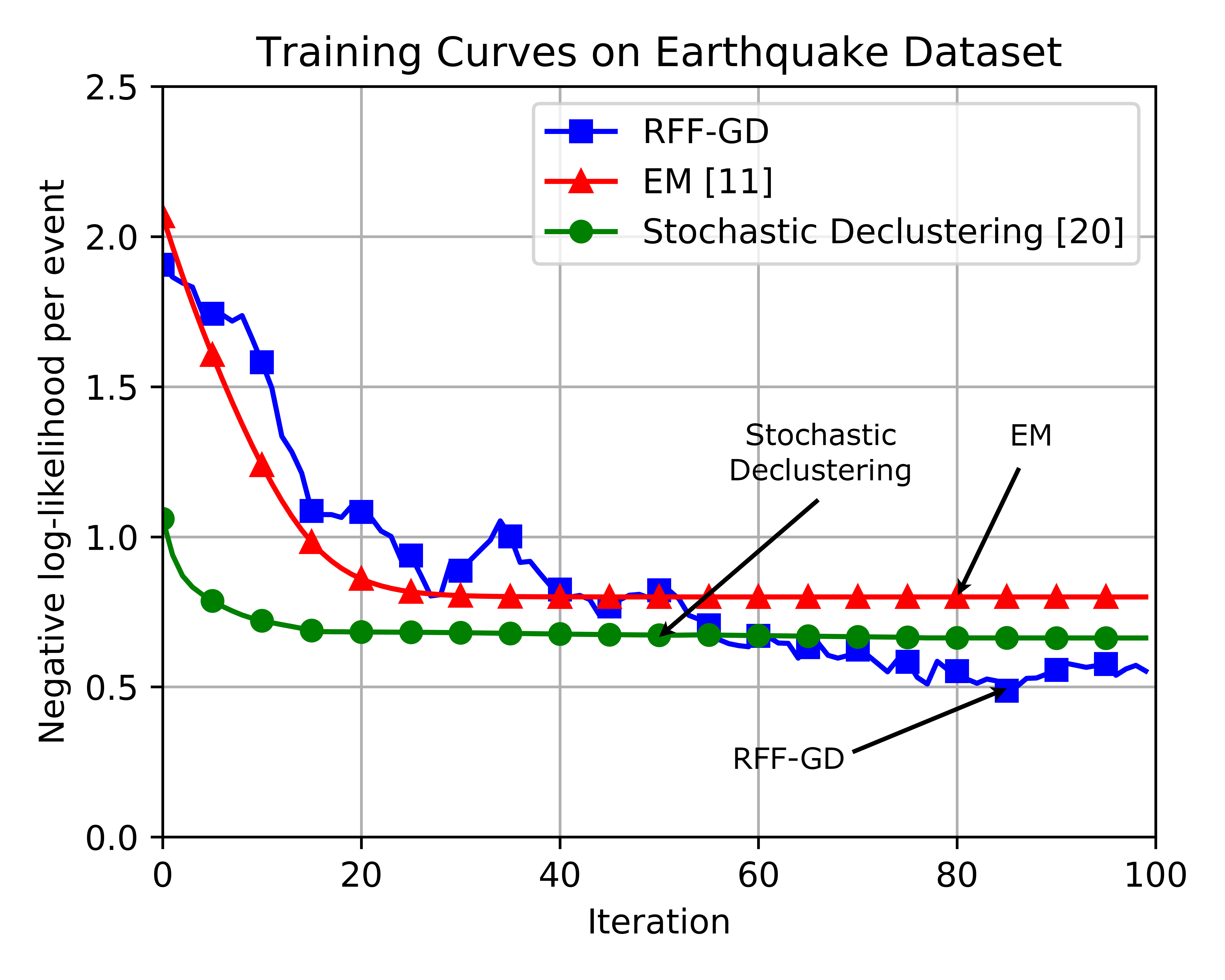}
\caption{} \label{fig1:r1}
\end{subfigure}
 \begin{subfigure}[t]{0.45\textwidth}
   \centering
\includegraphics[width=0.9\textwidth]{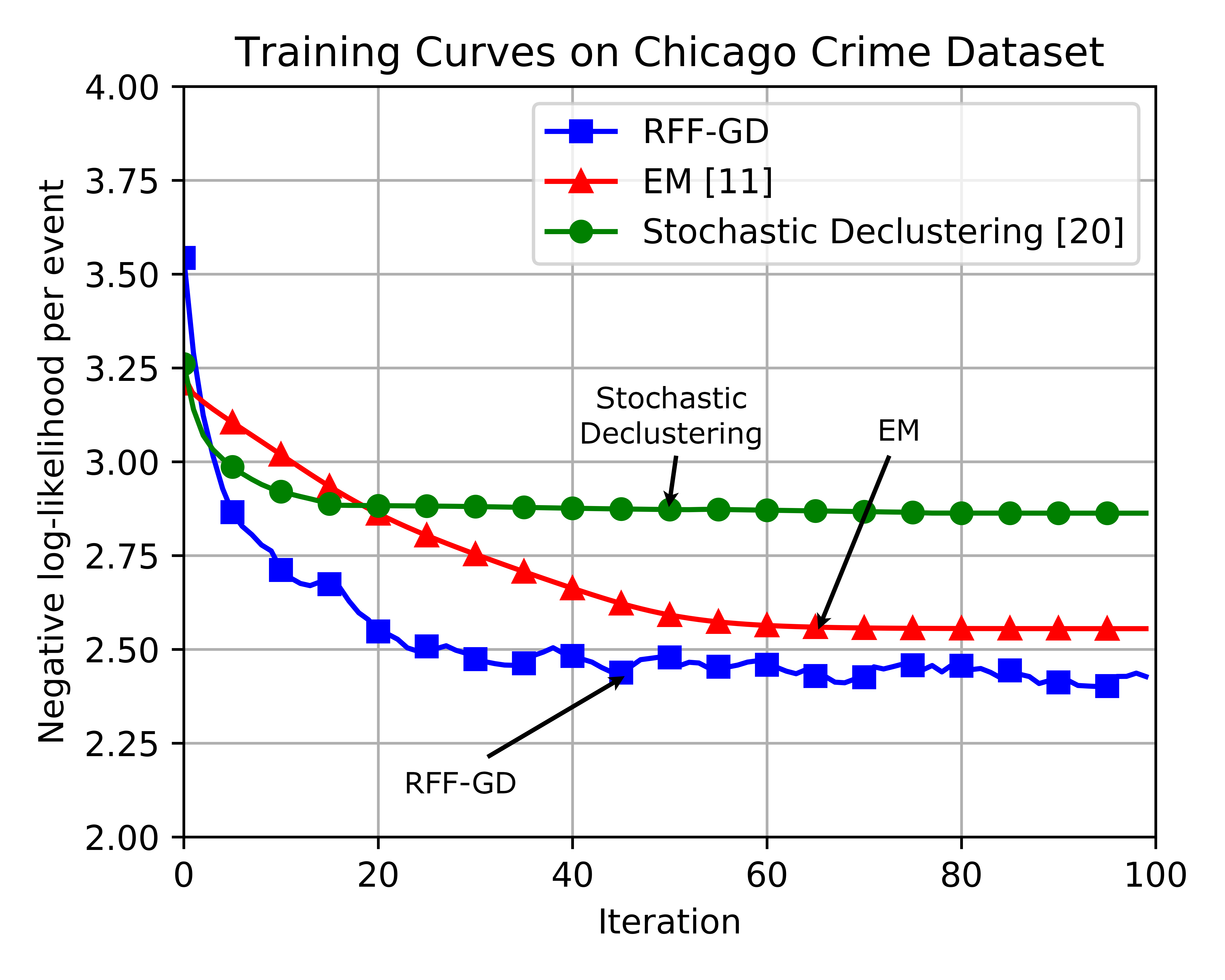}
\caption{} \label{fig1:r2}
\end{subfigure}
\caption{Training curves of the introduced method (RFF-GD), EM~\cite{yuan_mvhawkes} and stochastic declustering (SD)~\cite{zhuang_stodecl} for the earthquake dataset \textbf{(a)} and the Chicago crime dataset \textbf{(b)}.}\label{fig:1}
\end{figure*}
\noindent
\subsection{Real-life Dataset Experiments}

To investigate the fitting performance of our method in real-life datasets, we investigate the negative log-likelihood per event values and AIC as we have done in synthetic data. After learning the process parameters, we perform event analysis by examining the interactions between different event types in terms of excitation relations and spatio-temporal effects. We also investigate the effect of the number of dimensions in the randomized feature space. To this end, we have chosen the following two datasets. These datasets have been studied in the context of point processes, with applications on spatio-temporal prediction, and hotspot analysis~\cite{mohler_crime,ogata_earthquake}. They both exhibit certain characteristics such as having spatiotemporally clustered structures, which makes their modeling by spatio-temporal Hawkes processes plausible.

\subsubsection{Datasets}

\paragraph{Chicago Crime Dataset}
Chicago Crime Dataset includes the reported incidents in the City of Chicago from 2001, and is still being weekly updated by Chicago Police Department. The dataset includes the location and time of the incidents as well as their types such as theft, burglary, assault etc. Before collecting results, we grouped event types into four different classes considering their contextual meanings \textit{(1: Assault/Battery/Offense; 2: Burglary/Robbery/Theft; 3: Criminal Damage/Violations; 4:   Others)}. We particularly work in June 2019, and filter the locations spatially between the latitudes of [41.85, 41.92] and longitudes of [-87.65, -87.62] to remove outlier regions. 

\paragraph{Earthquake Dataset}
The National Earthquake Information Center provides this dataset that includes earthquakes with a magnitude of 4.5 or higher since 1986. Every earthquake entry includes a record of the date, time, location and magnitude. We filter the dataset spatially and work on the events occurred in Turkey, which is between the latitudes of [36, 42] and longitudes of [26, 45]. In addition, we have defined the event types according to the common categorization in the seismology literature on the Richter scale~\cite{richter} such that the first event type represents the earthquakes with a magnitude less than 5 (light), the second event type represents the earthquakes with a magnitude between 5 and 6 (moderate), and the third event type represents the earthquakes with a magnitude greater than 6 (strong). In the seismology literature, it has been shown that strong earthquakes cause aftershocks, i.e. earthquakes with small magnitudes~\cite{richter,ogata_earthquake,verejones}. Therefore, our representation enables us to infer the triggering relation between earthquakes from different magnitude ranges.

\subsubsection{Real-life Dataset Performance}

We investigate the fitting performance of the proposed optimization algorithm, and compare it with the EM algorithm proposed in a recent work~\cite{yuan_mvhawkes} and stochastic declustering~\cite{zhuang_stodecl}. As in the synthetic dataset experiments, we scale the given event sequence spatio-temporally. In addition, we repeat the experiments $10$ times to reduce the random effects on performance.

In the first set of experiments, we consider the earthquake dataset. We perform hyperparameter search over $D \in [10, 1000]$, $\eta \in [0.0001, 0.1]$, $b \in [32, 512]$ and $s \in [0.001, 0.1]$. We stop the training if the performance does not improve for $k=30$ consecutive steps and save the best iteration as our reference. For the proposed method, we obtain the best results for $D=60$, $\eta=0.002$, $b=512$ and $s=0.01$. For the second experiment set, we work on the Chicago crime dataset. We perform hyperparameter search over the same parameter ranges. In this experiment, we obtain the best performance with $D=40$, $\eta=0.01$, $b=256$ and $s=0.01$.
\renewcommand{\tabcolsep}{5pt}
\begin{table}[h]
\centering
\captionsetup{justification=centering}
\caption{\textsc{Training Performance of Our Algorithm, EM~\cite{yuan_mvhawkes} and Stochastic Declustering (SD)~\cite{zhuang_stodecl} on Real-life Datasets (\smaller{$p$: Number of parameters, $\overline{\mathcal{L}}$}: Negative log-likelihood per event)}}
\begin{tabular}{|c|c|c|c|c|c|c|}
\hline
\multirow{3}{*}{Method} & \multicolumn{6}{c|}{Dataset} \\ \cline{2-7} 
 & \multicolumn{3}{c|}{Earthquake} & \multicolumn{3}{c|}{Chicago} \\  \cline{2-7}
\rule{0pt}{8pt} & $p$ & $\overline{\mathcal{L}}$ & AIC & $p$ & $\overline{\mathcal{L}}$ & AIC \\ \hline
\textbf{RFF-GD} & $\boldsymbol{39}$ & $\boldsymbol{0.48}$ & $\boldsymbol{929.52}$ & $\boldsymbol{60}$ & $\boldsymbol{2.40}$ & $\boldsymbol{12480}$ \\ \hline
EM~\cite{yuan_mvhawkes} & $21$ & $0.80$ & $1461.2$ & $35$ & $2.55$ & $13202.5$ \\ \hline
SD~\cite{zhuang_stodecl} & $6$ & $0.66$ & $1182.84$ & $6$ & $2.86$ & $14741$ \\ \hline
\end{tabular}
\label{tab:fit_res}
\end{table}
\renewcommand{\tabcolsep}{6pt}
\begin{table}[h]
\centering
\captionsetup{justification=centering}
\caption{\textsc{Test Performance of Our Algorithm, EM~\cite{yuan_mvhawkes} and Stochastic Declustering (SD)~\cite{zhuang_stodecl} on Real-life Datasets in terms of Negative Log-likelihood per Event}}
\begin{tabular}{|c|c|c|}
\hline
\multirow{2}{*}{Method} & \multicolumn{2}{c|}{Dataset} \\ \cline{2-3} 
 & Earthquake & Chicago \\ \hline
\textbf{RFF-GD} & $\boldsymbol{0.83}$ & $\boldsymbol{2.51}$ \\ \hline
EM~\cite{yuan_mvhawkes} & $1.20$ & $2.68$ \\ \hline
SD~\cite{zhuang_stodecl} & $0.97$ & $2.90$ \\ \hline
\end{tabular}\label{tab:fit_res_test}
\end{table}
In Table ~\ref{tab:fit_res}, we provide the number of parameters, negative log-likelihood per event and AIC values of the EM algorithm~\cite{yuan_mvhawkes}, stochastic declustering~\cite{zhuang_stodecl}, and the introduced method for training set. Our method have more parameters due to the weight decay matrix and covariance matrices. We also provide the negative log-likelihood per event values for the test set in Table \ref{tab:fit_res_test}. On both datasets, our approach significantly outperforms other methods in terms of negative log-likelihood per event and AIC, which indicates that the inferred parameters by our method represent the given event sequence more successfully. We also illustrate the training curves for these experiments in Fig ~\ref{fig1:r1}, ~\ref{fig1:r2} respectively. 

\begin{figure}[htbp]
\begin{subfigure}[b]{0.45\columnwidth}
   \centering
\includegraphics[width=0.9\textwidth]{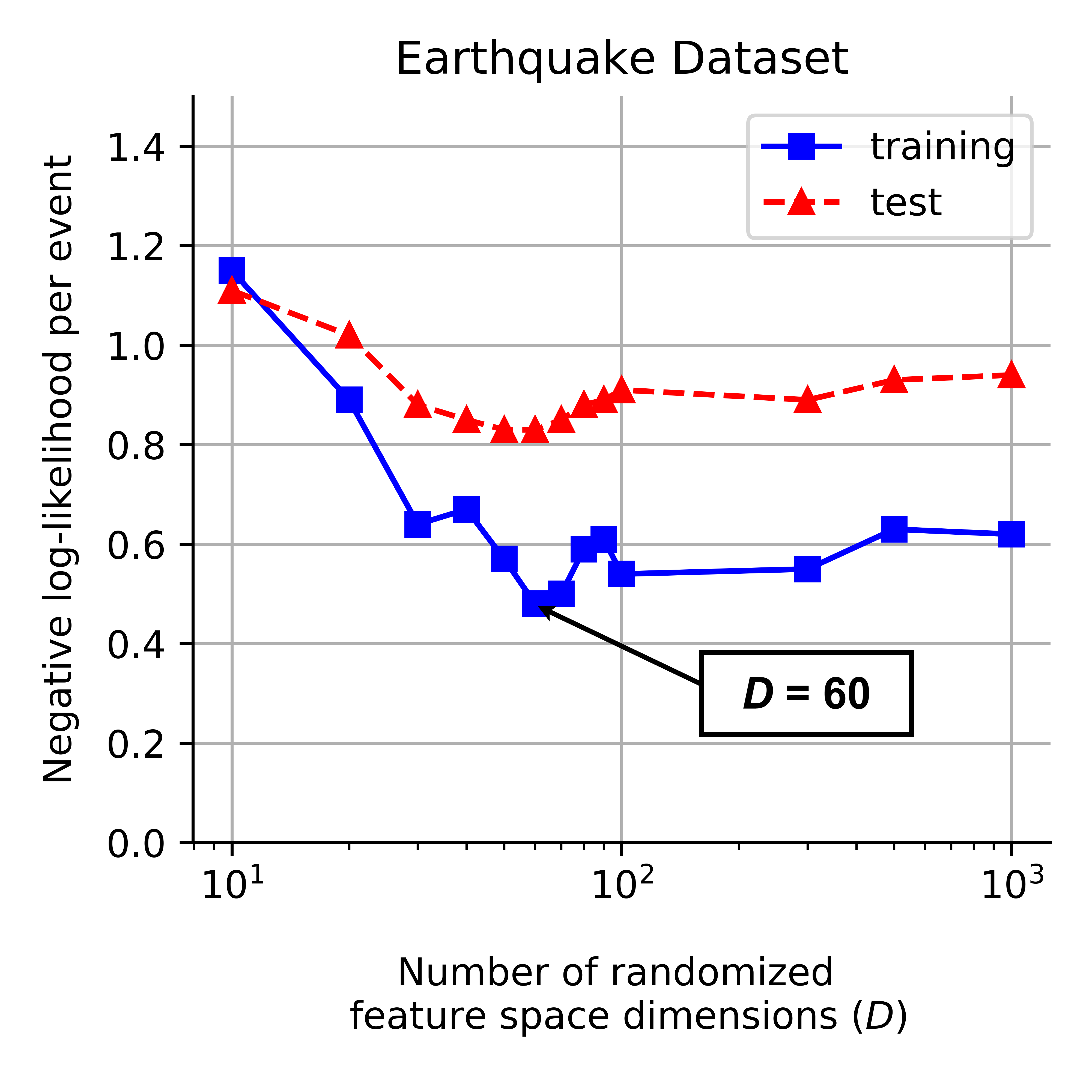}
\caption{} \label{fig5:d1}
\end{subfigure}\hfill
 \begin{subfigure}[b]{0.45\columnwidth}
   \centering
\includegraphics[width=0.9\textwidth]{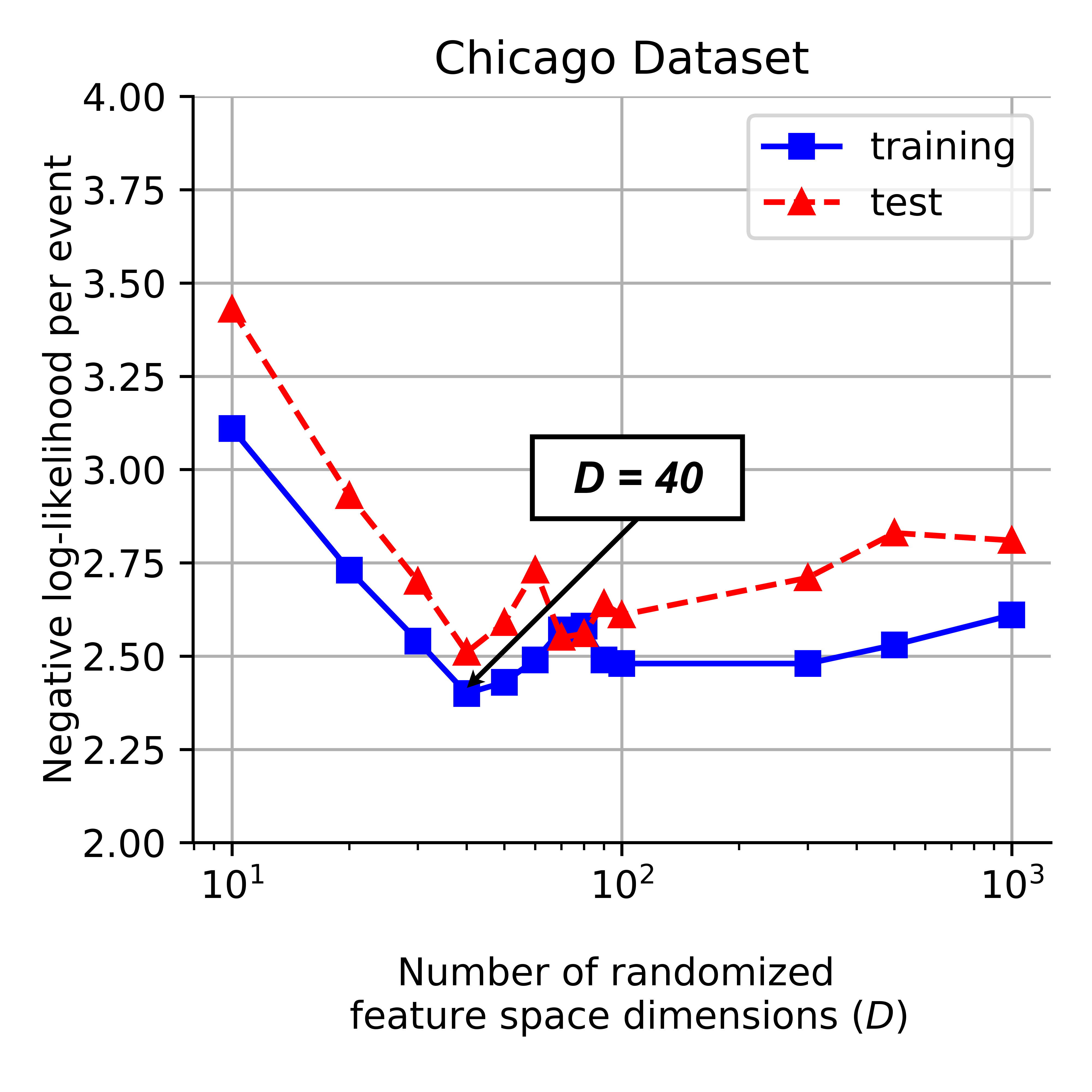}
\caption{} \label{fig5:d2}
\end{subfigure}
\caption{$D$ (number of randomized feature dimensions) vs. negative log-likelihood per event for the training and test sets of the earthquake dataset \textbf{(a)} and the Chicago crime dataset \textbf{(b)}.}
\label{fig5:d}  
\end{figure}

To illustrate the effect of the number of random Fourier feature dimensions, we provide Figure \ref{fig5:d}. In Figure \ref{fig5:d1}, for the earthquake dataset, we observe that the optimum choice for the randomized transformation dimensions is around 70. The performance significantly drops when $D$ becomes very small. If $D$ gets very high, we do not obtain a considerable amount of performance gain, in fact, the performance drops slightly. For the Chicago crime dataset, we observe a similar behavior as can be seen in Figure \ref{fig5:d2}. In this case, the best performance is achieved when $D=40$. Increasing this value causes negative log-likelihood per event to reach values between $2.5$ and $2.6$. Therefore, it is clear that the introduced tunable randomization while modeling the spatial excitation enhances the performance in real-life scenarios, particularly when the spatial dynamics of the underlying system deviates from pure Gaussian behavior.

\begin{figure}[h]
\begin{subfigure}[b]{0.5\columnwidth}
   \centering
\includegraphics[width=\linewidth]{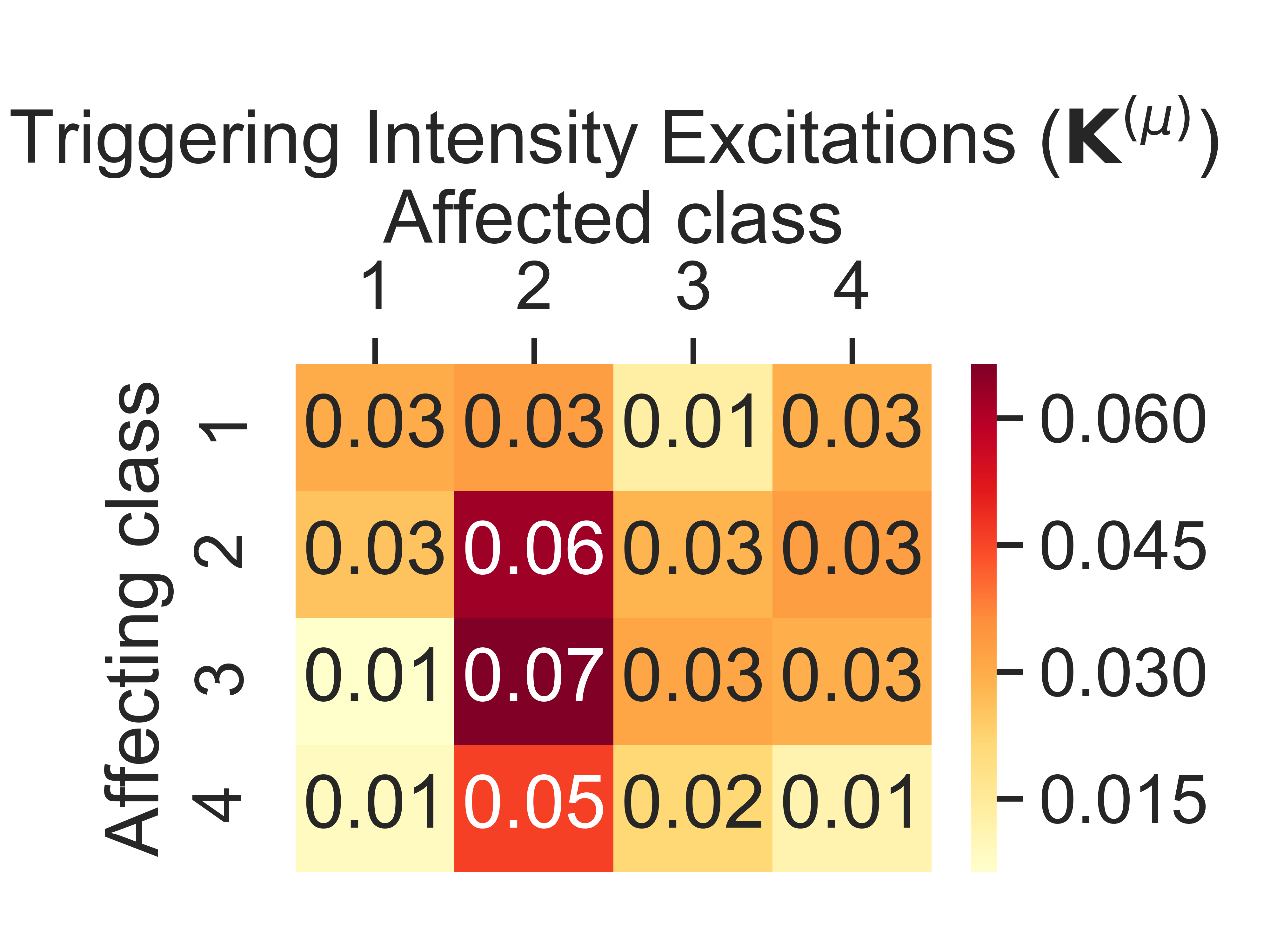}
\caption{} \label{fig1:r3}
\end{subfigure}\hfill
 \begin{subfigure}[b]{0.5\columnwidth}
   \centering
\includegraphics[width=\linewidth]{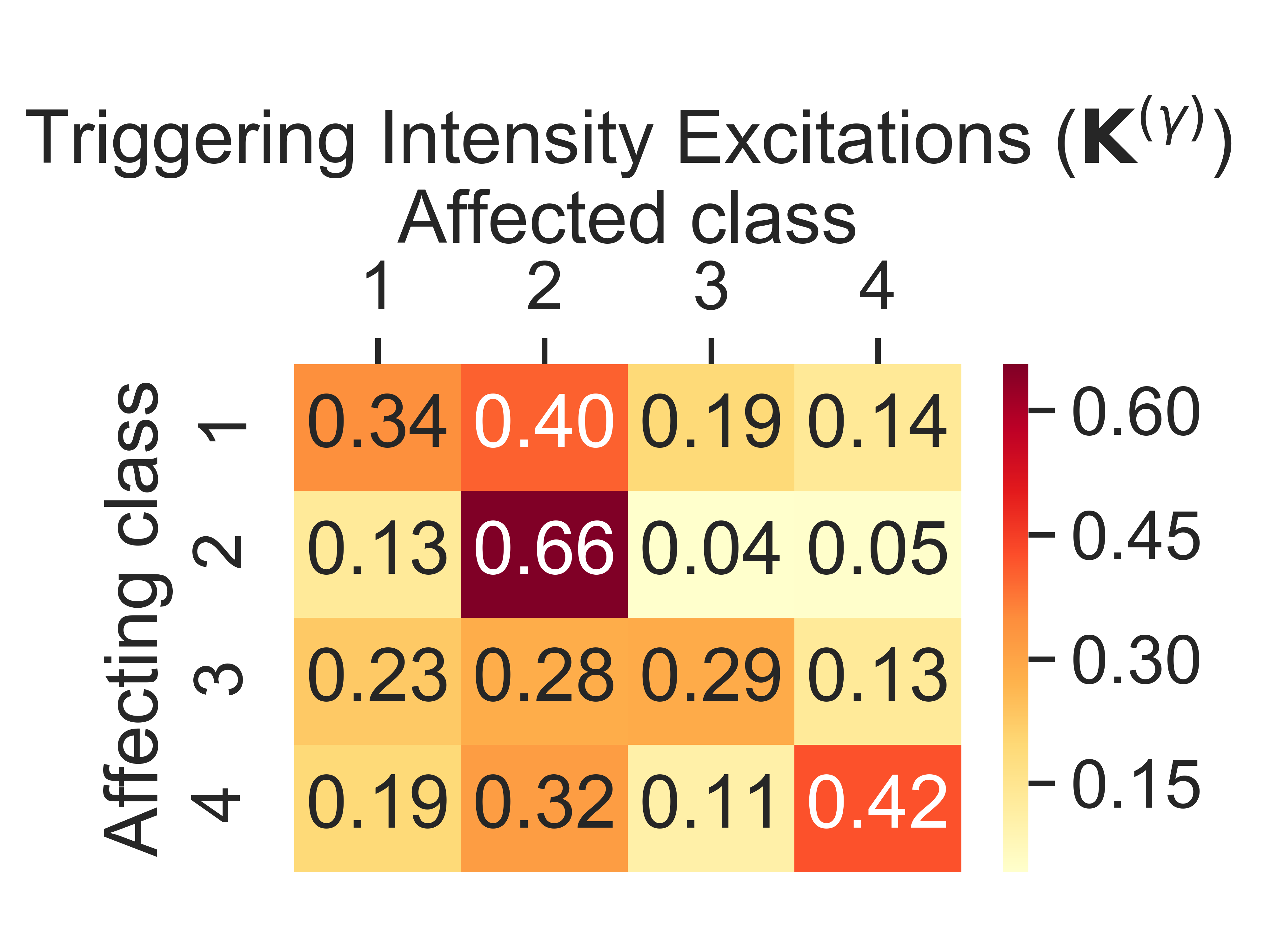}
\caption{} \label{fig1:r4}
\end{subfigure}
\caption{For the Chicago crime dataset, \textbf{(a)} Excitation matrix of base intensity ($\mathbf{K}^{(\mu)}$), \textbf{(b)} Excitation matrix of triggering intensity ($\mathbf{K}^{(\gamma)}$).}\label{fig:res_excitation}
\end{figure}

After investigating the fitting performance, we focus on the inferred parameters, which inherently reflect the dynamics of the given event sequence. For this purpose, we provide the estimated excitation matrices for the base and triggering intensities in Fig. \ref{fig:res_excitation}. This analysis directly reveals the triggering effect between different crime types. In this scenario, we observe that the base excitation values are more homogeneous compared to the triggering excitation values. In particular, crime events from class 2 (burglary/robbery/theft) have a strong self-excitation with respect to other event types. We also realize that events from class 2 are significantly triggered by other event types, whereas their effect on others is limited. On the contrary, events from class 3 (criminal damage/violations) exhibit strong excitation over all event types however, they are not considerably excited by other event types.

\section{Conclusion}
\label{sec:conclusion}
We studied spatio-temporal Hawkes processes to perform spatio-temporal event analysis. We introduce a novel framework for spatio-temporal Hawkes processes to extend the conventional methods in the literature such as EM~\cite{yuan_mvhawkes} and stochastic declustering~\cite{zhuang_stodecl}. Our approach utilizes the randomization introduced by random Fourier features based spatial kernel representation, and increases the flexibility of the model in terms of spatial modeling capability. Moreover, we express the problem in a neat scalable matrix formulation. We analytically calculate the intractable terms in the likelihood function, and derive the gradient equations for maximum likelihood optimization. To satisfy the structural constraints of the process parameters, we use reparameterization techniques and projected gradient descent. We also propose a thinning-based simulation algorithm for spatio-temporal Hawkes processes with multiple event types. We analyze the improvements achieved by the proposed method on various simulations and two real-life datasets. The comparisons show that the proposed method significantly performs better in terms of negative log-likelihood and AIC compared to other methods. In addition, we interpret the learned process parameters and perform event analysis over these real-life datasets through analyzing the triggering relations between event types.

\appendices
\section{}
\label{sec:app_a}
We can obtain the derivatives for the base and triggering intensity excitation matrices introduced in \eqref{eq:mu_intro} and \eqref{eq:gamma_intro} as
\noindent
\begin{align} \label{eq:deriv_start}
    & \frac{\partial\mathcal{L}}{\partial\mathbf{K^{(\cdot)}}} = - \mathbf{Q}^{{(\cdot)}^T}(\mathbf{Y} \oslash \mathbf{A}) + \frac{\partial R}{\partial\mathbf{K^{(\cdot)}}},
\end{align}
\noindent
where $\frac{\partial R}{\partial\mathbf{K^{(\cdot)}}}$ consists of the elements $\begin{bmatrix} \frac{\partial R}{\partial K^{(\cdot)}_{mn}} \end{bmatrix}$, which can be expressed as $ \frac{\partial R}{\partial K^{(\mu)}_{mn}} = \sum_{j=1}^N \delta_{u_j m} $ and $\frac{\partial R}{\partial K^{(\gamma)}_{mn}} = \sum_{j=1}^N \delta_{u_j m} (1-e^{-w_{u_j n}(T-t_j)})$. We, then, express the derivative of the decay rate matrix as $\frac{\partial\mathcal{L}}{\partial\mathbf{W}} =
    \begin{bmatrix}
        \frac{\partial\mathcal{L}}{\partial w_{mn}} \\
    \end{bmatrix}$, 
where each element is derived as
\noindent
\begin{align}    
    & \frac{\partial\mathcal{L}}{\partial w_{mn}} = -\text{sum}((\frac{\partial\mathbf{A}}{\partial w_{mn}})^T(\mathbf{Y} \oslash \mathbf{A})) + \frac{\partial R}{\partial w_{mn}}.
\end{align}
\noindent
Here, $ \frac{\partial R}{\partial w_{mn}} = \sum_{j=1}^N \delta_{u_j m}(T-t_j)e^{-w_{u_j n}(T-t_j)}$, 
\noindent
and $\frac{\partial\mathbf{A}}{\partial w_{mn}}$ consists of the rows
\noindent
\begin{align}
     & \frac{\partial \boldsymbol{a}_i^T}{\partial w_{mn}} =  \frac{1}{2\pi}|\mathbf{\Sigma^{(\gamma)}}|^{-1/2}\boldsymbol{z}_i^T\mathbf{Z}_{J(t_i)}^{(\gamma)^T}\mathrm{diag}(\frac{\partial \boldsymbol{d}_{J(t_i)}}{\partial w_{mn}})\mathbf{Y}_{J(t_i)}\mathbf{K}^{(\gamma)},
\end{align}
\noindent
where
\noindent
\begin{align}
    & \frac{\partial \boldsymbol{d}_{J(t_i)}}{\partial w_{mn}} = 
    \begin{bmatrix}
        \vdots\\
        \delta_{u_i m}\delta_{u_j n}(1-w_{u_j u_i}(t_i - t_j))e^{-w_{u_j u_i}(t_i-t_j)}\\
        \vdots
    \end{bmatrix}.
\end{align}
\noindent

For the spatial kernel parameters, we first derive the gradients of $\mathbf{V}^{(\mu)}$ and $\mathbf{V}^{(\gamma)}$ as $\frac{\partial\mathcal{L}}{\partial\mathbf{V}^{(\cdot)}} =
    \begin{bmatrix}
        \frac{\partial\mathcal{L}}{\partial V_{mn}^{(\cdot)}},
    \end{bmatrix}$
where each element is derived as
\noindent
\begin{align}
    & \frac{\partial\mathcal{L}}{\partial V_{mn}^{(\cdot)}} = -\text{sum}\left((\frac{\partial\mathbf{A}}{\partial V_{mn}^{(\cdot)}})^T(\mathbf{Y} \oslash \mathbf{A})\right).
\end{align}
\noindent
Here, $\frac{\partial\mathbf{A}}{\partial V_{mn}^{(\cdot)}}$ consists of the rows $\frac{\partial \boldsymbol{a}_i^T}{\partial V_{mn}^{(\cdot)}}$ such that
\noindent
\begin{align}
    & \frac{\partial \boldsymbol{a}_i^T}{\partial V_{mn}^{(\mu)}} = \frac{1}{2\pi T}|\mathbf{\Sigma}^{(\mu)}|^{-1/2}(\frac{\partial \boldsymbol{z_i}^{(\mu)^T}}{\partial V_{mn}^{(\mu)}}\mathbf{Z}_{J(t_i)}^{(\mu)^T}  \nonumber \\ & \hspace{.39in} + \boldsymbol{z_i}^{(\mu)^T}\frac{\partial{\mathbf{Z}_{J(t_i)}^{(\mu)^T}}}{\partial V_{mn}^{(\mu)}})\mathbf{Y}_{J_{(t_i)}} \mathbf{K}^{(\mu)}, \\
    & \frac{\partial \boldsymbol{a}_i^T}{\partial V_{mn}^{(\gamma)}} = \frac{1}{2\pi}|\mathbf{\Sigma}^{(\gamma)}|^{-1/2}(\frac{\partial \boldsymbol{z_i}^{(\gamma)^T}}{\partial V_{mn}^{(\gamma)}}\mathbf{Z}_{J(t_i)}^{(\gamma)^T} \nonumber \\ & \hspace{.39in} + \boldsymbol{z_i}^{(\gamma)^T}\frac{\partial{\mathbf{Z}_{J(t_i)}^{(\gamma)^T}}}{\partial V_{mn}^{(\gamma)}})\mathrm{diag}(\boldsymbol{d}_{J(t_i)})\mathbf{Y}_{J(t_i)} \mathbf{K}^{(\gamma)},
\end{align}
\noindent
where $\frac{\partial{\mathbf{Z}_{J(t_i)}^{{(\cdot)}^T}}}{\partial V_{mn}^{(\cdot)}}$ consists of the rows $\frac{\partial \boldsymbol{z_j}^{{(\cdot)}^T}}{\partial V_{mn}^{(\cdot)}}=-\sqrt{\frac{2}{ D}}s_{im}\ell_n^{1/2}\boldsymbol{u_n}^T\odot\sin{(\boldsymbol{s_i}^T\mathbf{V}^{(\cdot)}\mathbf{\Lambda^{{(\cdot)}^{-1/2}}}\mathbf{U} + \boldsymbol{b}^T)}$, and $\boldsymbol{u_n}^T$ is the $n^{th}$ row of $\mathbf{U}$. Finally, we obtain the derivatives of $\boldsymbol{\ell}^{(\mu)}$ and $\boldsymbol{\ell}^{(\gamma)}$ as $\frac{\partial\mathcal{L}}{\partial\boldsymbol{\ell}} =
    {\begin{bmatrix}
        \hdots & \frac{\partial\mathcal{L}}{\partial \ell_n} & \hdots \\
    \end{bmatrix}}^T$, where each element is defined as $\frac{\partial\mathcal{L}}{\partial \ell_n} = -\text{sum}\left((\frac{\partial\mathbf{A}}{\partial \ell_n})^T(\mathbf{Y} \oslash \mathbf{A})\right)$. Here, $\frac{\partial\mathbf{A}}{\partial \ell_{n}}$ consists of the rows $\frac{\partial \boldsymbol{a}_i^T}{\partial \ell_n}$ such that
\noindent
\begin{align}
    & \frac{\partial \boldsymbol{a}_i^T}{\partial \ell_n^{(\mu)}} = \frac{1}{2\pi T}(\frac{\partial |\mathbf{\Sigma}^{(\mu)}|^{-1/2}}{\partial \ell_n}\boldsymbol{z_i}^{(\mu)^T}\mathbf{Z}_{J(t_i)}^{(\mu)^T} \nonumber \\ & \hspace{.34in} + |\mathbf{\Sigma}^{(\mu)}|^{-1/2}(\frac{\partial \boldsymbol{z_i}^{(\mu)^T}}{\partial \ell_n}\mathbf{Z}_{J(t_i)}^{(\mu)^T}  \nonumber \\ & \hspace{.34in} + \boldsymbol{z_i}^{(\mu)^T}\frac{\partial{\mathbf{Z}_{J(t_i)}^{(\mu)^T}}}{\partial \ell_n}))\mathbf{Y}_{J(t_i)} \mathbf{K}^{(\mu)}, \\
    & \frac{\partial \boldsymbol{a}_i^T}{\partial \ell_n^{(\gamma)}} = \frac{1}{2\pi}(\frac{\partial |\mathbf{\Sigma}^{(\gamma)}|^{-1/2}}{\partial \ell_n}\boldsymbol{z_i}^{(\gamma)^T}\mathbf{Z}_{J(t_i)}^{(\gamma)^T} \nonumber \\ & \hspace{.34in} + |\mathbf{\Sigma}^{(\gamma)}|^{-1/2}(\frac{\partial \boldsymbol{z_i}^{(\gamma)^T}}{\partial \ell_n}\mathbf{Z}_{J(t_i)}^{(\gamma)^T} \nonumber \\ & \hspace{.34in} + \boldsymbol{z_i}^{(\gamma)^T}\frac{\partial{\mathbf{Z}_{J(t_i)}^{(\gamma)^T}}}{\partial \ell_n}))\mathrm{diag}(\boldsymbol{d}_{J(t_i)})\mathbf{Y}_{J(t_i)} \mathbf{K}^{(\gamma)},
    \label{eq:deriv_end} 
\end{align}
\noindent
where $\frac{\partial{\mathbf{Z}_{J(t_i)}^{{(\cdot)}^T}}}{\partial \ell_{n}^{(\cdot)}}$ consists of the rows $\frac{\partial \boldsymbol{z_j}^{{(\cdot)}^T}}{\partial \ell_{n}^{(\cdot)}}=\sqrt{\frac{1}{ 2D}} \boldsymbol{s}_i^T\boldsymbol{v}_n^{(\cdot)}\ell_n^{{(\cdot)}^{-1/2}}\boldsymbol{u_n}^T\odot\sin{(\boldsymbol{s_i}^T\mathbf{V}^{(\cdot)}\mathbf{\Lambda^{{(\cdot)}^{-1/2}}}\mathbf{U} + \boldsymbol{b}^T)}$, $\boldsymbol{v_n}^{(\cdot)}$ is the $n^{th}$ column of $\mathbf{V}^{(\cdot)}$ and $\boldsymbol{u_n}^T$ is the $n^{th}$ row of $\mathbf{U}$.

\bibliographystyle{IEEEtran}
\balance
\bibliography{main}

\end{document}